\documentclass[10pt,twocolumn,letterpaper]{article}

\usepackage[pagenumbers]{cvpr} 

\usepackage{graphicx}
\usepackage{amsmath}
\usepackage{amssymb}
\usepackage{booktabs}
\usepackage{multirow}

\usepackage{color,xcolor}
\usepackage{epsfig}
\usepackage{graphicx}
\usepackage{times}

\usepackage{comment}

\usepackage{pifont}

\usepackage{microtype}
\usepackage[font=small]{caption}
\usepackage{arydshln}
\usepackage{tabularx}
\usepackage{adjustbox}
\usepackage{array}
\usepackage{booktabs}
\usepackage{colortbl}
\usepackage{float,wrapfig}
\usepackage{hhline}
\usepackage{multirow}
\usepackage{subcaption} %
\usepackage[percent]{overpic}

\usepackage{stmaryrd}
\usepackage{breqn}

\usepackage{amsmath,amsfonts,amssymb}

\usepackage{bm}
\usepackage{nicefrac}
\usepackage{microtype}
\usepackage{dsfont}
\usepackage{changepage}
\usepackage{extramarks}
\usepackage{fancyhdr}
\usepackage{setspace}
\usepackage{soul}
\usepackage{xspace}
\usepackage{hhline}
\usepackage{algorithmicx}
\usepackage{algorithm}
\usepackage{algpseudocode}
\usepackage{pifont}
\usepackage{amssymb}

\usepackage[pagebackref,breaklinks,colorlinks]{hyperref}

\usepackage{enumitem}
\usepackage[title]{appendix}

\newcommand{\cmark}{\ding{51}}%
\newcommand{\xmark}{\ding{55}}

\def\X{\mathcal{X}} %
\def\P{\mathbb{P}} %
\def\Px{\mathbb{P}_{\mathcal{X}}} %
\def\Pg{\mathbb{P}_{\theta}} %

\def\D{{D}}
\def\G{{G}}

\newcommand{\reffig}[1]{Figure~\ref{fig:#1}}
\newcommand{\refsec}[1]{Section~\ref{sec:#1}}
\newcommand{\refapp}[1]{Appendix~\ref{sec:#1}}
\newcommand{\reftbl}[1]{Table~\ref{tbl:#1}}
\newcommand{\refalg}[1]{Algorithm~\ref{alg:#1}}

\newcommand{\refeq}[1]{Eqn.~\ref{eq:#1}}

\newcommand{\lblfig}[1]{\label{fig:#1}}
\newcommand{\lblsec}[1]{\label{sec:#1}}
\newcommand{\lbleq}[1]{\label{eq:#1}}
\newcommand{\lbltbl}[1]{\label{tbl:#1}}
\newcommand{\lblalg}[1]{\label{alg:#1}}

\newcommand{\ignorethis}[1]{}

\newcommand{\myparagraph}[1]{\vspace{1pt} \noindent \textbf{#1} \ }

\def\1{\bm{1}}

\newcolumntype{L}[1]{>{\raggedright\let\newline\\\arraybackslash\hspace{0pt}}m{#1}}
\newcolumntype{C}[1]{>{\centering\let\newline\\\arraybackslash\hspace{0pt}}m{#1}}
\newcolumntype{R}[1]{>{\raggedleft\let\newline\\\arraybackslash\hspace{0pt}}m{#1}}

\newcommand{\ignore}[1]{}

\renewcommand*{\thefootnote}{\arabic{footnote}}

\makeatletter
\DeclareRobustCommand\onedot{\futurelet\@let@token\@onedot}
\def\@onedot{\ifx\@let@token.\else.\null\fi\xspace}

\makeatother

\usepackage[capitalize]{cleveref}
\crefname{section}{Sec.}{Secs.}
\Crefname{section}{Section}{Sections}
\Crefname{table}{Table}{Tables}
\crefname{table}{Tab.}{Tabs.}

\begin{document}

\title{Ensembling Off-the-shelf Models for GAN Training}

\author{Nupur Kumari \textsuperscript{1}
\qquad
Richard Zhang\textsuperscript{2}
\qquad
Eli Shechtman\textsuperscript{2}
\qquad
Jun-Yan Zhu\textsuperscript{1}\\
\textsuperscript{1}Carnegie Mellon University
\qquad
\textsuperscript{2}Adobe
}

\maketitle

\begin{abstract}
The advent of large-scale training has produced a cornucopia of powerful visual recognition models. However, generative models, such as GANs, have traditionally been trained from scratch in an unsupervised manner. Can the collective ``knowledge'' from a large bank of pretrained vision models be leveraged to improve GAN training? If so, with so many models to choose from, which one(s) should be selected, and in what manner are they most effective? We find that pretrained computer vision models can significantly improve performance when used in an ensemble of discriminators. Notably, the particular subset of selected models greatly affects performance. We propose an effective selection mechanism, by probing the linear separability between real and fake samples in pretrained model embeddings, choosing the most accurate model, and progressively adding it to the discriminator ensemble. Interestingly, our method can improve GAN training in both limited data and large-scale settings. Given only 10k training samples, our \textsc{FID} on \textsc{LSUN Cat} matches the StyleGAN2 trained on 1.6M images. On the full dataset, our method improves \textsc{FID} by $1.5$ to $2\times$ on cat, church, and horse categories of \textsc{LSUN}. 
\end{abstract}
\vspace{-12pt}
\section{Introduction}
\lblsec{intro}

Image generation inherently requires being able to capture and model complex statistics in real-world visual phenomena. Computer vision models, driven by the success of supervised and self-supervised learning techniques~\cite{vgg,resnet, chen2020simple, mocoimproved, radford2021learning}, have proven effective at capturing useful representations when trained on large-scale data~\cite{russakovsky2015imagenet, yu2015lsun, zhou2017places}. What potential implications does this have on generative modeling? If one day, perfect computer vision systems could answer any question about any image, could this capability be leveraged to improve image synthesis models?

\begin{figure}[t]
    \centering
    \includegraphics[width=0.4\textwidth]{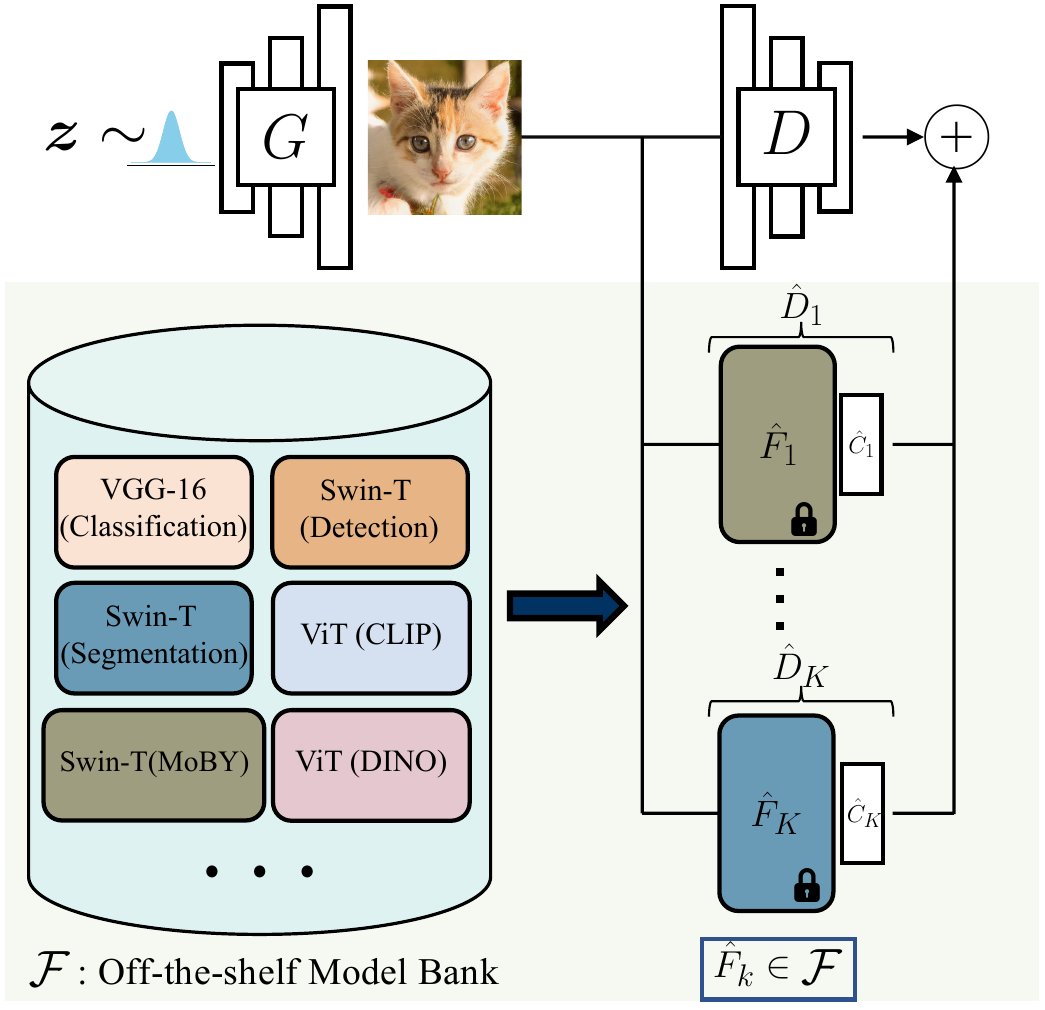}
    \caption{{\textbf{Vision-aided GAN training}. 
    The model bank $\mathcal{F}$ consists of widely used and state-of-the-art pretrained networks. We automatically select a subset $\{\hat{F}\}_{k=1}^K$ from $\mathcal{F}$, which can best distinguish between real and fake distribution. Our training procedure consists of creating an ensemble of the original discriminator $\D$ and discriminators $\hat{D}_k = \hat{C}_k \circ \hat{F}_k$ based on the feature space of selected off-the-shelf models. $\hat{C}_k$ is a shallow trainable network over the frozen pretrained features.}}
    \lblfig{methodology}
    \vspace{-15pt}
\end{figure}

Surprisingly, despite the aforementioned connection between synthesis and analysis, state-of-the-art generative adversarial networks (GANs)~\cite{stylegan3, diffaug, biggan, stylegan2ada} are trained in an unsupervised manner without the aid of such pretrained networks. With a plethora of useful models easily available in the research ecosystem, this presents a missed opportunity to explore. Can the knowledge of pretrained visual representations actually benefit GAN training?  If so, with so many models, tasks, and datasets to choose from, which models should be used, and in what manner are they most effective?

In this work, we study the use of a ``bank'' of pretrained deep feature extractors to aid in generative model training. Specifically, GANs are trained with a discriminator, aimed at continuously learning the relevant statistics differentiating real and generated samples, and a generator, which aims to reduce this gap.
Na\"ively using such strong, pretrained networks as a discriminator leads to the overfitting and overwhelming the generator, especially in limited data settings. We show that freezing the pretrained network (with a small, lightweight learned classifier on top, as shown in \reffig{methodology}) provides stable training when used with the original, learned discriminator.
In addition, ensembling multiple pretrained networks encourages the generator to match the real distribution in different, complementary feature spaces.

To choose which networks work best, we propose to use an automatic model selection strategy based on the linear separability of real and fake images in the feature space, and progressively add supervision from a set of available pretrained networks. In addition, we use label smoothing~\cite{improvedgan_labelsmoothing} and differentiable augmentation~\cite{diffaug,stylegan2ada} to stabilize the model training further and reduce overfitting.

We experiment on several datasets in both limited and large-scale sample setting to show the effectiveness of our method. We improve the state-of-the-art on \textsc{FFHQ}~\cite{karras2019style} and \textsc{LSUN}~\cite{yu2015lsun} datasets given 1k training samples by 2-$3\times$ on the FID metric~\cite{fid}. For \textsc{LSUN Cat}s, we match the FID of StyleGAN2 trained on the full dataset (1.6M images) with only 10k samples, as shown in \reffig{1ktofull}. In the full-scale data setting, our method improves FID for \textsc{LSUN Cat}s from $6.86$ to $3.98$, \textsc{LSUN Church} from $4.28$ to $1.72$, and \textsc{LSUN Horse} from $4.09$ to $2.11$. Finally, we visualize the internal representation of our learned models as well as training dynamics. Our \href{https://github.com/nupurkmr9/vision-aided-gan}{code} is available on our \href{https://www.cs.cmu.edu/~vision-aided-gan/}{website}. 


\section{Related Work}\lblsec{related}

\myparagraph{Improving GAN training.} Since the introduction of GANs~\cite{goodfellow2020generative}, significant advances have been induced by architectural changes~\cite{radford2015unsupervised,karras2019style,stylegan3}, training schemes~\cite{karras2018progressive,zhang2017stackgan},  as well as objective functions~\cite{arjovsky2017WGAN,mao2017least,mescheder2018R1,albuquerque2019multi,durugkar2016generative,doan2019line}. In previous works, the learning objectives often aim to minimize different types of divergences between real and fake distribution. The discriminators are typically trained from scratch and do not use external pretrained networks. As a result, the discriminator is prone to overfit the training set, especially for the limited data setting~\cite{diffaug,stylegan2ada,yang2021data}.

\myparagraph{Use of pretrained models in image synthesis.}
Pretrained models have been widely used as perceptual loss functions~\cite{dosovitskiy2016generating,gatys2016image,johnson2016perceptual} to measure the distance between an output image and a target image in deep feature space. The loss has proven effective for conditional image synthesis tasks such as super-resolution~\cite{ledig2016photo}, image-to-image translation~\cite{chen2017photographic,wang2018pix2pixHD,park2019semantic}, and neural style transfer~\cite{gatys2016image}. Zhang et al.~\cite{zhang2018unreasonable} show that deep features can indeed match the human perception of image similarity better than classic metrics. Sungatullina et al.~\cite{perceptual_disc} propose a perceptual discriminator to combine perceptual loss and adversarial loss for unpaired image-to-image translation. This idea was recently used by a concurrent work on CG2real~\cite{richter2021enhancing}. Another recent work~\cite{gadde2021detail} proposes the use of pretrained objects detectors to detect regions in the image and train object-specific discriminators during GAN training. Our work is inspired by the idea of perceptual discriminators~\cite{perceptual_disc} but differs in three ways. First, we focus on a different application of unconditional GAN training rather than image-to-image translation. Second, instead of using a single VGG model, we ensemble a diverse set of feature representations that complement each other. Finally, we propose an automatic model selection method to find models useful for a given domain. A concurrent work~\cite{sauer2021projected} propose to reduce overfitting of perceptual discriminators~\cite{perceptual_disc} using random projection and achieve better and faster GAN training.

Loosely related to our work, other works have used pretrained models for clustering, encoding, and nearest neighbor search during their model training.  Logo-GAN~\cite{sage2018logo} uses deep features to get synthetic clustering labels for conditional GAN training.  InclusiveGAN~\cite{yu2020inclusive} improves the recall of generated samples by enforcing each real image to be close to a generated image in deep feature space. Shocher et al.~\cite{shocher2020semantic} uses an encoder-decoder based generative model with pretrained encoder for image-to-image translation tasks. Pretrained features have also been used to condition the generator in GANs~\cite{mangla2020data,casanova2021instance}. Different from the above work, our method empowers the discriminator with pretrained models and requires no changes to the backbone generator.

\myparagraph{Use of pretrained models in image editing.}
Pretrained models have also been used in image editing once the generative model has been trained. Notable examples include image projection with a perceptual distance~\cite{zhu2016generative,abdal2019image2stylegan}, text-driven image editing with CLIP~\cite{patashnik2021styleclip}, finding editable directions using attribute classifier models~\cite{shen2020interfacegan}, and extracting semantic editing regions with pretrained segmentation networks~\cite{zhu2021barbershop}. In our work, we focus on using the rich knowledge of computer vision models to improve model training. 

\myparagraph{Transfer learning.}
Large-scale supervised and self-supervised models learn useful feature representations~\cite{krizhevsky2012imagenet,he2016deep,chen2020simple,xie2021self,caron2021emerging,radford2021learning} that can transfer well to unseen tasks, datasets, and domains~\cite{donahue2014decaf,zeiler2014visualizing,yosinski2014transferable,saenko2010adapting,tzeng2017adversarial,zamir2018taskonomy,oquab2014learning,huh2016makes,robusttransferimagenet}. 
In generative modeling, recent works propose transferring the weights of pretrained generators and discriminators from a  source domain (e.g., faces) to a new domain (e.g., portraits of one person)~\cite{wang2018TransferGAN,noguchi2019SB,wang2019MineGAN,mo2020FreezeD,zhao2020leveraging,ojha2021few,liu2020towards,gu2021lofgan}. Together with differentiable data augmentation techniques~\cite{stylegan2ada,zhao2020image,tran2020towards,diffaug}, they have shown faster convergence speed and better sampling quality for limited-data settings.
Different from them, we transfer the knowledge of learned feature representations of computer vision models. This enables us to leverage the knowledge from a diverse set of sources at scale. 

\section{Method}
\lblsec{method}

\begin{figure*}[!t]
    \centering
    \includegraphics[width=0.9\textwidth]{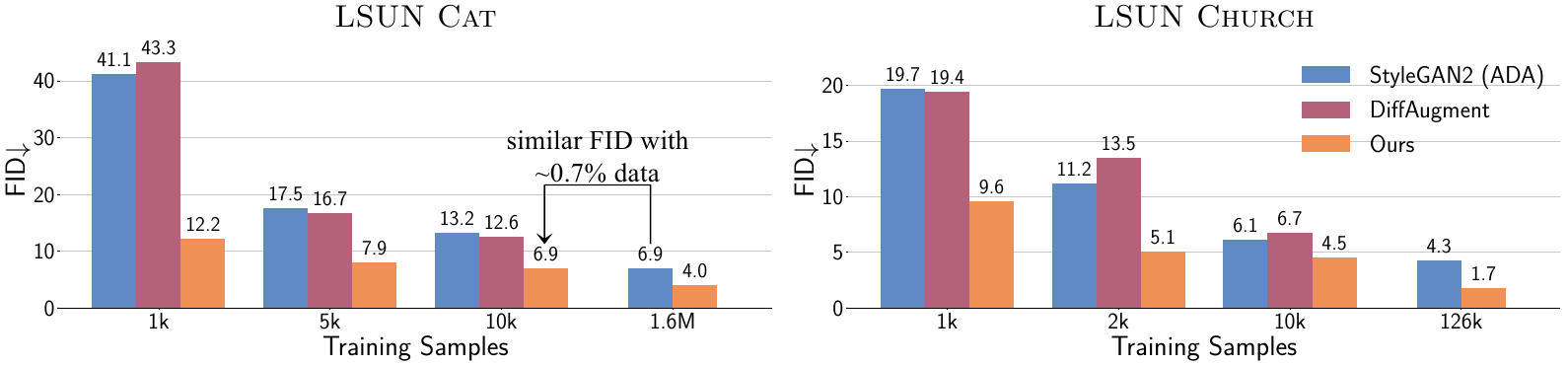}
     \vspace{-6pt}
    \caption{\textbf{Performance on \textsc{LSUN Cat} and \textsc{LSUN Church}}. We compare with the leading methods StyleGAN2-ADA~\cite{stylegan2ada} and DiffAugment~\cite{diffaug} on different sizes of training samples and full-dataset. Our method outperforms them by a large margin, especially in limited sample setting. For \textsc{LSUN Cat} we achieve similar FID as StyleGAN2~\cite{stylegan2} trained on full-dataset using only $0.7\%$ of the dataset.
    }
    \lblfig{1ktofull}
    \vspace{-8pt}
\end{figure*}

Generative Adversarial Networks (GANs) aim to approximate the distribution of real samples from a finite training set $\bm x \sim \Px$. The generator network $\G$, maps latent vectors $\bm z \sim \P(\bm z)$ (e.g., a normal distribution) to samples $\G(\bm z) \sim \Pg$. The discriminator network $\D$ is trained adversarially to distinguish between the continuously changing generated distribution $\P_{\theta}$ and target real distribution $\Px$. GANs perform the minimax optimization $\min_{G} \max_{\D}  V(D, G)$, where
\begin{equation}
\begin{gathered}
    V(D, G)  = \mathbb{E}_{\bm x\sim \Px } [\log \D(\bm x)] + \mathbb{E}_{\bm z\sim \P(\bm z) } [ \log(1 - \D(\G(\bm z))) ].
\end{gathered}
\lbleq{gans}
\end{equation}

Ideally, the discriminator should measure the gap between $\Px$ and $\Pg$ and guide the generator towards $\Px$. However, in practice, large capacity discriminators can easily overfit on a given training set, especially in the limited-data regime~\cite{diffaug,stylegan2ada}. Unfortunately, as shown in \reffig{overfit}, even when we adopt the latest differentiable data augmentation~\cite{stylegan2ada} to reduce overfitting, the discriminator still tends to overfit, failing to perform well on a validation set. In addition, the discriminator can potentially focus on artifacts that are indiscernible to humans but obvious for machines~\cite{wang2020cnn}.

\begin{figure}[t]
    \centering
    \includegraphics[width=\linewidth]{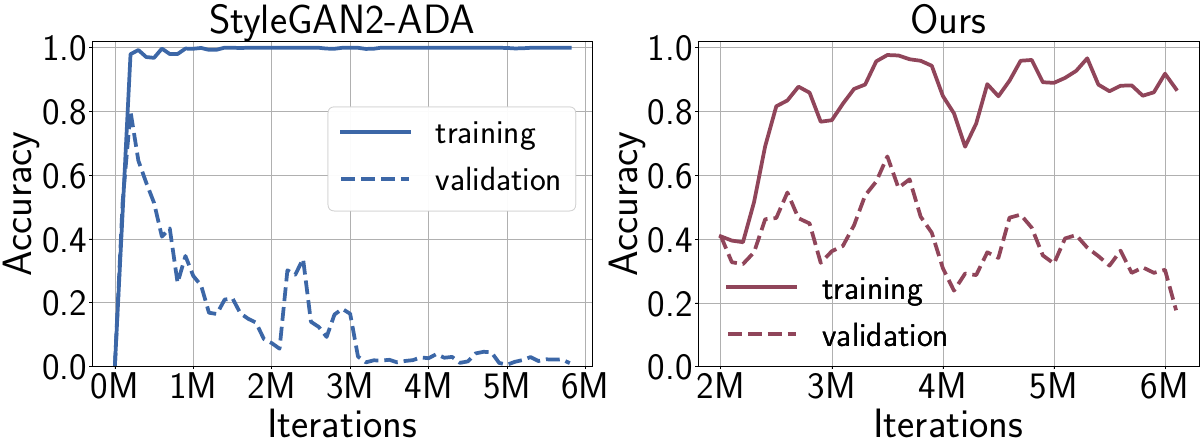}
     \vspace{-8pt}
    \caption{Training and validation accuracy w.r.t. training iterations for our DINO~\cite{caron2021emerging} based discriminator vs. baseline StyleGAN2-ADA discriminator on FFHQ 1k dataset. Our discriminator based on pretrained features has higher accuracy on validation real images and thus shows better generalization. In the above training, vision-aided adversarial loss is added at the 2M iteration.
    }
    \lblfig{overfit}
    \vspace{-8pt}
\end{figure}

To address the above issues, we propose ensembling a diverse set of deep feature representations as our discriminator. This new source of supervision can benefit us in two manners. First, training a shallow classifier over pretrained features is a common way to adapt deep networks to a small-scale dataset, while reducing overfitting~\cite{girshick2014rich,chen2019closer}. 
As shown in \reffig{overfit}, our method reduces the discriminator overfitting significantly. Second, recent studies~\cite{zeiler2014visualizing,bau2020understanding} have shown that deep networks can capture meaningful visual concepts from low-level visual cues (edges and textures) to high-level concepts (objects and object parts). A discriminator built on these features may better match human perception~\cite{zhang2018unreasonable}. 

\subsection{Formulation}
Given a set of pretrained feature extractors $\mathcal{F} = \{F_n\}_{n=1}^{N}$, which learns to tackle different vision tasks, we train corresponding discriminators $\{D_{n}\}_{n=1}^{N}$. We add small classifier heads $\{C_n\}_{n=1}^N$ to measure the gap between $\Px$ and $\Pg$ in the pretrained models' feature spaces. During discriminator training, the feature extractor $F_n$ is frozen, and only the classifier head is updated. The generator $\G$ is updated with the gradients from $\D$ and the discriminators $\{D_{n}\}$ based on pretrained feature extractors. In this manner, we propose to leverage pretrained models in an adversarial fashion for GAN training, which we refer to as \textit{Vision-aided Adversarial} training:
\begin{equation}
\begin{aligned}
    & \min_{G}\; \max_{\D ,\{C_n\}_{n=1}^{N}} V(D, G) + \overbrace{ \sum_{n=1}^{N} V(D_n,G)}^{\text{vision-aided adversarial loss} },\\
    & \textnormal{where  } D_n = C_n \circ F_n.
    \lbleq{baseline}
\end{aligned}
\end{equation}

Here, $C_n$ is a small trainable head over the pretrained features. The above training objective involves the sum of discriminator losses based on all available pretrained models $\{F_n\}$. Solving for this at each training iteration would be computationally and memory-intensive. Using all pretrained models would force a significant reduction in batch size to fit all models into memory, potentially hurting performance~\cite{biggan}.  
To bypass the computational bottleneck, we automatically select a small subset of $K$ models, where $K < N$: 
\begin{equation}
\begin{aligned}
  \min_{G}\; \max_{\D ,\{\hat{C}_{k}\}_{k=1}^{K}} V(D, G) 
  + \sum_{k=1}^{K} V(\hat{D}_{k},G),
    \lbleq{ours}
\end{aligned}
\end{equation}
where $\hat{D}_k=\hat{C}_k \circ \hat{F}_k$ denotes the discriminator corresponding to $k^{\text{th}}$ selected model, and $k \in \{1, \dots, K\}$.

\begin{algorithm}[t] 
\small
\caption{\small{GAN training with \emph{Vision-aided Adversarial} loss.}}\lblalg{gan_algo}
\begin{algorithmic}[1]
\renewcommand{\algorithmicrequire}{\textbf{Input:}}
\renewcommand{\algorithmicensure}{\textbf{Output:}}
\algnewcommand\algorithmicinput{\textbf{Hyperparameters:}}
\algnewcommand\Hyperparameters{\item[\algorithmicinput]}
    \Require  $\G$, $\D$ trained with standard GAN loss for baseline number of iterations. Off-the-shelf model bank $\mathcal{F} = \{F_n\}_{n=1}^N$. Training data $\{\bm x_i \}$.
    \Hyperparameters $K$: maximum number of pretrained models to use. $\{T_k : k =1\cdots K\}$: training intervals before adding next pretrained model. 
    \State Selected model set $\mathcal{\hat{F}} = $\O
    
    \For{ $k= 1$ to $K$} 
    \State  Select best model $\hat{F}_k \in \mathcal{F}$ using \refeq{4}
    \State $\mathcal{\hat{F}}= \mathcal{\hat{F}} \cup \{ \hat{F}_k \}$
    \State $\hat{D}_k = \hat{C}_k \circ \Hat{F}_k $ 
    \Comment $\hat{C}_k$ is a shallow trainable network
    \State $\mathcal{F}= \mathcal{F}\setminus \hat{F}_k$
    
    \For{ $t= 1$ to $T_k$}
    \State Sample $\bm x \sim \{\bm x_i \} $ 
    \State Sample $ \bm z \sim  \P(\bm z)$ 
    \State Update $\D, \hat{D}_j $ $\forall j=1,\cdots, k$ using \refeq{ours}   
    \State Sample $ \bm z \sim  \P(\bm z)$
    \State Update $\G$ using \refeq{ours} 
    \EndFor
    \EndFor
    \Ensure $\G$ with best training set FID 
\end{algorithmic}
\end{algorithm}

\subsection{Model Selection}\lblsec{3.2}
We choose the models whose off-the-shelf feature spaces best distinguish samples from real and fake distributions. Given the pretrained model's features of real and fake images, the strongest adversary from the set of models is $\hat{F}_{k}$, where
\begin{equation}
    \begin{aligned}
    & k = \arg \max_{n} \big \{ \max_{C'_n} V(D'_n, G)\big \}, \\
    & \text{where } D'_n = C'_n \circ F_n. 
    \lbleq{4}
    \end{aligned}
\end{equation}

Here $F_n$ is frozen, and $C'_n$ is a linear trainable head over the pretrained features. In the case of limited real samples available and for computational efficiency, we use linear probing to measure the separability of real and fake images in the feature space of $F_n$.

We split the union of real training samples $\{ \bm x_i \}$ and generated images $\{\G(\bm z_i)\}$ into training and validation sets. For each pretrained model $F_n$, we train a logistic linear discriminator head to classify whether a sample comes from $\P_{\X}$ or $ \P_{\theta}$ and measure $V(D'_n, G)$ on the validation split. The above term measures the negative binary cross-entropy loss and returns the model with the lowest error. A low validation error correlates with higher accuracy of the linear probe, indicating that the features are useful for distinguishing real from generated samples and using these features will provide more useful feedback to the generator. We empirically validate this on GAN training with 1k training samples of \textsc{FFHQ} and \textsc{LSUN Cat} datasets. \reffig{linear_acc} shows that the GANs trained with the pretrained model $F_n$ with higher linear probe accuracy in general achieve better FID metrics.

To incorporate feedback from multiple off-the-shelf models, we explore two variants of model selection and ensembling strategies -- (1) \textbf{K-fixed} model selection strategy chooses the K best off-the-shelf models at the start of training and trains until convergence and (2) \textbf{K-progressive} model selection strategy iteratively selects and adds the best, unused off-the-shelf model after a fixed number of iterations.

\myparagraph{K-progressive model selection. }
We find including multiple models in a progressive manner has lower computational complexity compared to the K-fixed strategy. This also helps in the selection of pretrained models, which captures different aspects of the data distribution. For example, the first two models selected through the progressive strategy are usually a pair of self-supervised and supervised models. For these reasons, we primarily perform all of our experiments using the progressive strategy. We also show a comparison between the two strategies in \refsec{ablation}.

\begin{figure}[t]
    \centering
    \includegraphics[width=0.4\textwidth]{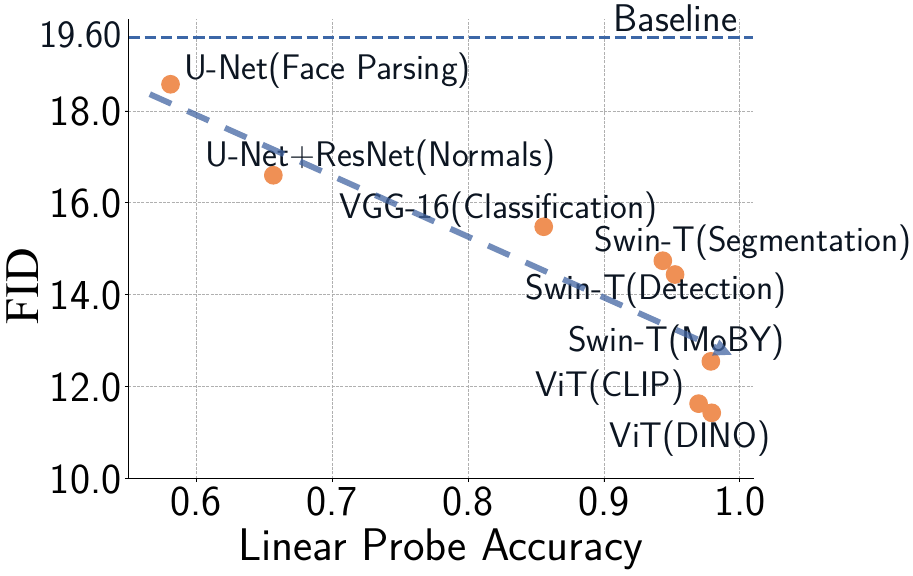}
    \vspace{-8pt}
    \caption{\textbf{Model selection using linear probing of pretrained features}. We show correlation of FID with the accuracy of a logistic linear model trained for real vs fake classification over the features of off-the-shelf models. Top dotted line is the FID of StyleGAN2-ADA generator used in model selection and from which we finetune with our proposed vision-aided adversarial loss. Similar analysis for \textsc{LSUN Cat} is shown in \reffig{fig:appendix-lsuncat-correlation-fid} in the appendix. } 
    \lblfig{linear_acc}
    \vspace{-12pt}
\end{figure}

\myparagraph{Discussion.} 
The idea of linear separability as a metric has been previously used for evaluating GAN via classifier two-sample tests~\cite{twosampletestclassifier,zhao2021large}. We adopt this in our work to evaluate the usefulness of available off-the-shelf discriminators, rather than evaluating generators. ``Linear probing'' is also a common technique for measuring the effectiveness of intermediate features spaces in both self-supervised~\cite{zhang2017split,He_2020_CVPR,chen2020simple} and supervised~\cite{alain2016understanding} contexts, and model selection has been explored in previous works to predict expert models for transfer learning~\cite{puigcerver2020scalable,mustafa2020deep,dwivedi2019representation}. We explore this in context of generative modeling and propose a progressive addition of next best model to create an ensemble~\cite{caruana2004ensemble} of discriminators. 

\begin{table*}[!t]
\centering
\setlength{\tabcolsep}{5pt}
\scalebox{.8}{
\begin{tabular}{ @{\extracolsep{0pt}} l@{\hspace{0.2\tabcolsep}}r @{\hspace{0.2\tabcolsep}}r@{\hspace{0.98\tabcolsep}} r  r  c c c c c c  @{}}
\toprule
\multicolumn{2}{l}{\multirow{2}{*}{\textbf{Dataset}}}
& \multirow{2}{*}{\textbf{StyleGAN2}}
& \multirow{2}{*}{\textbf{DiffAugment}}
& \multirow{2}{*}{\textbf{ADA}}
&\multicolumn{3}{c}{\textbf{ Ours (w/ ADA)} } 
&\multicolumn{3}{c}{\textbf{ Ours (w/ DiffAugment)} } \\
\cmidrule{6-8} \cmidrule{9-11} 
& & & & & \textbf{+$1^{\text{st}}$ D} &\textbf{+$2^{\text{nd}}$ D} &\textbf{+$3^{\text{rd}}$ D} & \textbf{+$1^{\text{st}}$ D} &\textbf{+$2^{\text{nd}}$ D} &\textbf{+$3^{\text{rd}}$ D}  \\
\midrule
\multirow{3}{*}{\rotatebox{90}{\textbf{FFHQ}}} 
& 1k   & 62.16 & 27.20 & 19.57  & 11.43 & \textbf{10.39} & 10.58 & \textbf{12.33} & 13.39 & 12.76 \\ 
& 2k   & 42.62 & 16.63 & 16.06  & 10.17 & 8.73 & \textbf{8.18}& 10.01 & \textbf{9.24}  & 10.99 \\ 
& 10k  & 16.07 & 8.15  &  8.38 & 6.90  & 6.39 & \textbf{5.90} & 6.94  & \textbf{6.26} & 6.43 \\ 
\midrule
\multirow{3}{*}{\rotatebox{90}{\begin{tabular}{@{}c@{}} \textbf{LSUN}\\ \textbf{\textsc{Cat}} \end{tabular} }} 
& 1k  & 185.75 & 43.32 & 41.14  &  15.49 & 12.90 & \textbf{12.19}  & 13.52 & 12.52  & \textbf{11.01}\\ 
& 2k   & 68.03  & 25.70 & 23.32  & 13.44 &  13.35  & \textbf{11.51} & 12.20  & 11.79 & \textbf{11.33} \\ 
& 10k  & 18.59 & 12.56 & 13.25  & 8.37 &  7.13  & \textbf{6.86} & 8.19   & 7.90 &  \textbf{7.79} \\ 
\midrule
\multirow{3}{*}{\rotatebox{90}{ \begin{tabular}{@{}c@{}} \textbf{LSUN}\\ \textbf{\textsc{Church}} \end{tabular} }} 
& 1k   & - & 19.38 & 19.66 &  11.39 & 9.78 & \textbf{9.56} & 10.15  & \textbf{9.87} & 9.94\\ 
& 2k   & - & 13.46 & 11.17 & 5.25 & \textbf{5.06}    &  5.26 & 6.09 & 6.37  & \textbf{5.56} \\ 
& 10k  & - & 6.69 & 6.12 & 4.80  & 4.82  &  \textbf{4.47}  & 3.42 & 3.41 & \textbf{3.25} \\\vspace{-5pt}
\\ 
\bottomrule
\end{tabular}}
\vspace{-5pt}
\caption{\textbf{FFHQ and \textsc{LSUN} results} with varying training samples from 1k to 10k. 
FID$\downarrow$ is measured with complete dataset as reference distribution. We select the best snapshot according to training set FID, and report mean of 3 FID evaluations. In Ours (w/ ADA) we finetune the StyleGAN2-ADA model, and in Ours (w/ DiffAugment) we finetune the model trained with DiffAgument while using the corresponding policy for augmentation. Our method works with both ADA and DiffAugment strategy for augmenting images input to the discriminators.
}

\lbltbl{few-shot-training-vary}
\vspace{-15pt}
\end{table*}

\subsection{Training Algorithm} 
\lblsec{training}
As shown in \refalg{gan_algo}, our final algorithm consists of first training a GAN with standard adversarial loss~\cite{stylegan2, goodfellow2020generative}. Given this baseline generator, we search for the best off-the-shelf models using linear probing and introduce our proposed loss objective during training. In the K-progressive strategy, we add the next vision-aided discriminator after training for a fixed number of iterations proportional to the number of available real training samples. The new vision-aided discriminator is added to the snapshot with the best training set FID in the previous stage. During training, we perform data augmentation through horizontal flipping and use differentiable augmentation techniques~\cite{stylegan2ada,diffaug} and one-sided label smoothing~\cite{improvedgan_labelsmoothing} as a regularization. We also observe that only using off-the-shelf models as the discriminator leads to divergence. Thus, the benefit is brought by ensembling the original discriminator and the newly added off-the-shelf models. We show results with the use of three pretrained models and observe minimal benefit with the progressive addition of next model if the linear probe accuracy is low and worse than the models already in the selected set.

\section{Experiments}

\lblsec{expr}
 Here we conduct extensive experiments on multiple datasets of different resolutions with the StyleGAN2 architecture. We show results on \textsc{FFHQ}~\cite{karras2019style}, \textsc{LSUN Cat}, and \textsc{LSUN Church} datasets~\cite{yu2015lsun} while varying training sample size from 1k to 10k, as well as with the full dataset. For real-world limited sample datasets, we perform experiments on the cat, dog, and wild categories of AFHQ~\cite{choi2020stargan} dataset at 512 resolution and \textsc{MetFaces}~\cite{stylegan2ada} at 1024 resolution. In 100-400 low-shot settings, we perform experiments on AnimalFace cat and dog~\cite{si2011learning}, and 100-shot Bridge-of-Sighs~\cite{diffaug} dataset. We also show results with BigGAN~\cite{biggan} architecture on CIFAR~\cite{krizhevsky2009learning} datasets in \refapp{appendix-2}.

\myparagraph{Baseline and metrics.} We compare with state-of-the-art methods for limited dataset GAN training, StyleGAN2-ADA~\cite{stylegan2ada} and DiffAugment~\cite{diffaug}. We compute the commonly used Fréchet Inception Distance (FID) metric~\cite{fid} using the \texttt{clean-fid} library~\cite{cleanfid} to evaluate models. In low-shot settings we evaluate on KID~\cite{binkowski2018demystifying} metric as well. We report more evaluation metrics like precision and recall~\cite{improvedprecisionrecall}, and SwAV-FID~\cite{morozov2020self,kynkaanniemi2022role} using feature space of SwAV~\cite{caron2020unsupervised} model which was not used during our training in \refapp{appendix-3}.

\myparagraph{Off-the-shelf models.} We include eight large-scale self-supervised and supervised networks. Specifically, we perform experiments with CLIP~\cite{radford2021learning}, VGG-16~\cite{vgg} trained for ImageNet~\cite{deng2009imagenet} classification, and self-supervised models, DINO~\cite{caron2021emerging} and MoBY~\cite{xie2021self}. We also include face parsing~\cite{lee2020maskgan} and face normals prediction networks~\cite{facenormal}. Finally, we have Swin-Transformer~\cite{liu2021swin} based segmentation model trained on ADE-20K~\cite{zhou2019semantic} and object detection model trained on MS-COCO~\cite{lin2014microsoft}. Full details of all models is given in \reftbl{cv_models} in \refapp{appendix-4}. We exclude the Inception model~\cite{szegedy2016rethinking} trained on ImageNet since Inception features have already been used to calculate the FID metric.

\begin{figure}[!t]
    \centering
    \includegraphics[width=0.43\textwidth]{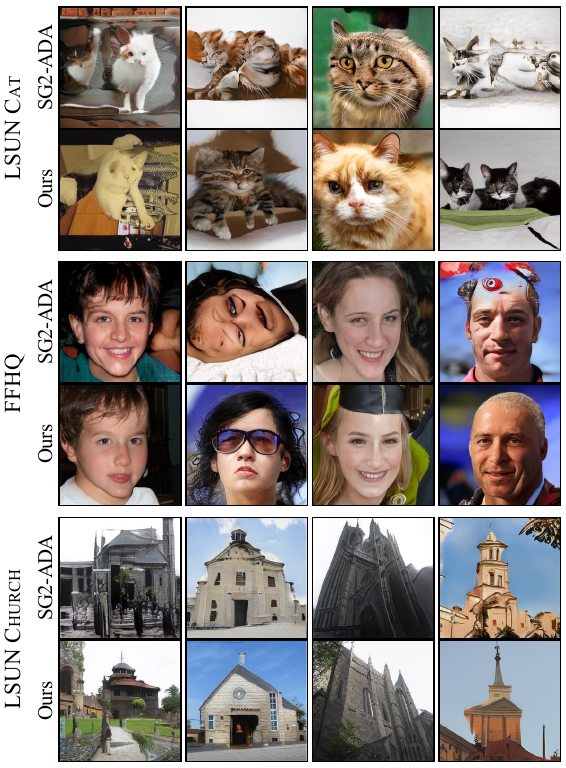}
    \vspace{-10pt}
    \caption{\textbf{\textsc{LSUN Cat}, \textsc{FFHQ}, and \textsc{LSUN Church} paired sample comparison in 1k training dataset setting}. For each dataset, the top row shows the baseline StyleGAN2-ADA samples, and the bottom row shows the samples by Our method for the same randomly sample latent code. We fine-tune the StyleGAN2-ADA model with our vision-aided adversarial loss. For the same latent code image quality improves with our method on average. 
    }
    \lblfig{train_vary_samples} 
    \vspace{-8pt}
\end{figure}

\myparagraph{Vision-aided discriminator's architecture.}
For discriminator $\hat{D}_k$ based on pretrained model features, we extract spatial features from the last layer and use a small  \texttt{Conv-LeakyReLU-Linear-LeakyReLU-Linear} architecture for binary classification. In the case of big transformer networks, such as CLIP and DINO, we explore a multi-scale architecture that works better. For all experiments, we use three pretrained models selected by the model selection strategy during training. Details about the architecture, model training, memory requirements, and hyperparameters are provided in \refapp{appendix-4}.
\begin{figure}[t]
    \centering
    \includegraphics[width=0.48\textwidth]{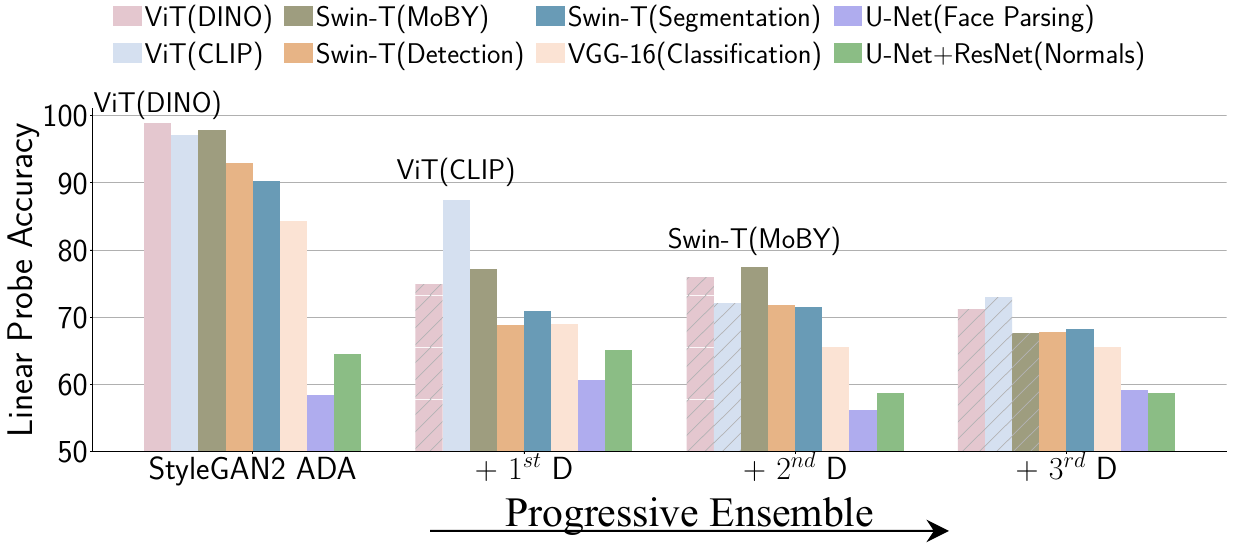}
    \caption{\textbf{Linear probe accuracy of off-the-shelf models during our K-progressive ensemble training} on FFHQ 1k. For the StyleGAN2-ADA, ViT (DINO) model has the highest accuracy and is selected first, then ViT (CLIP) and then Swin-T (MoBY). As we train with vision-aided discriminators, linear probe accuracy decreases for most of the pretrained models. Similar trend for all our experiments are shown in the \refapp{appendix-4}.}
    \lblfig{linear_acc_ffhq}
    \vspace{-10pt}
\end{figure}

\subsection{\textbf{\textsc{FFHQ}} and \textbf{\textsc{LSUN}} datasets}
\reftbl{few-shot-training-vary} shows the results of our method when the training sample is varied from 1k to 10k for \textsc{FFHQ}, \textsc{LSUN Cat}, and \textsc{LSUN Church} datasets. The considerable gain in FID for all settings shows the effectiveness of our method in the limited data scenario. To qualitatively analyze the difference between our method and StyleGAN2-ADA, we show randomly generated samples from both models given the same latent code in \reffig{train_vary_samples}. Our method improves the quality of the worst samples, especially for \textsc{FFHQ} and \textsc{LSUN Cat} (also see \reffig{fig:appendix-train_vary_samples}, \ref{fig:appendix-train_vary_samples_diffaug} in \refapp{appendix-1}). \reffig{linear_acc_ffhq} shows the accuracy of linear probe over the pretrained models's features as we progressively add the next discriminator. 

To analyze the overfitting behavior of discriminators, we evaluate its training and validation accuracy across iterations. Compared to the baseline StyleGAN2-ADA discriminator, our vision-aided discriminator shows better generalization on the validation set specifically for limited-data regime as shown in \reffig{overfit} for \textsc{FFHQ} 1k setting.

\myparagraph{Full-dataset training.} In the full-dataset setting, we fine-tune the trained StyleGAN2 (config-F)~\cite{stylegan2} model with our method. \reftbl{full-dataset} shows the comparison of StyleGAN2 and ADM~\cite{dhariwal2021diffusion} with our method trained using three vision-aided discriminators. We report both FID and Perceptual Path Length (PPL)~\cite{karras2019style} (W space) metric. On \textsc{LSUN Cat}, our method improves FID from $6.86$ to $3.98$, on \textsc{LSUN Church} from $4.28$ to $1.72$, and on \textsc{LSUN Horse} from $4.09$ to $2.11$. For \textsc{FFHQ} dataset, our method improves the PPL metric from $144.62$ to $127.58$ and has similar performance on FID metric. Perceptual path length has been shown to correlate with image quality and indicates a smooth mapping in generator latent space~\cite{stylegan2}. Random generated samples for all models are shown in \reffig{fig:appendix-random_ffhq_lsun_full} in \refapp{appendix-1}.

\begin{table}[t]
\centering
\scalebox{0.8}{
\begin{tabular}{@{\extracolsep{4pt}}llcccc@{}}
\toprule
\multirow{2}{*}{\textbf{Dataset}}
&\multicolumn{2}{c}{\textbf{StyleGAN2 (F)}} 
&\multicolumn{2}{c}{\textbf{Ours (w/ ADA)}} &\multirow{1}{*}{\textbf{ADM}} \\
\cmidrule{2-3} \cmidrule{4-5} \cmidrule{6-6} 
 & FID $\downarrow$ & PPL $\downarrow$ & FID $\downarrow$ & PPL $\downarrow$ & FID $\downarrow$ \\
\midrule
\textsc{FFHQ-1024}  & \textbf{2.98}  & 144.62  & 3.01 & \textbf{127.58} & - \\ 
\textsc{LSUN Cat-256}  & 6.86 & 437.13 & \textbf{3.98} &  \textbf{420.15} & 5.57$^{*}$  \\ 
\textsc{LSUN Church-256} & 4.28 & \textbf{343.02} & \textbf{1.72} & 388.94 & - \\ 
\textsc{LSUN Horse-256} & 4.09 & 337.98 & \textbf{2.11} &  \textbf{307.12}  &  2.57$^{*}$ \\ 
\bottomrule
\end{tabular}}
\caption{\textbf{Results on full-dataset setting}. we improve the FID metric on \textsc{LSUN} categories by a significant margin. On the \textsc{FFHQ} dataset we improve the PPL metric. $*$ means directly reported from the ADM paper~\cite{dhariwal2021diffusion}.}
\lbltbl{full-dataset}
  \vspace{-10pt}
\end{table}

\begin{figure}[t]
    \centering
    \includegraphics[width=0.43\textwidth]{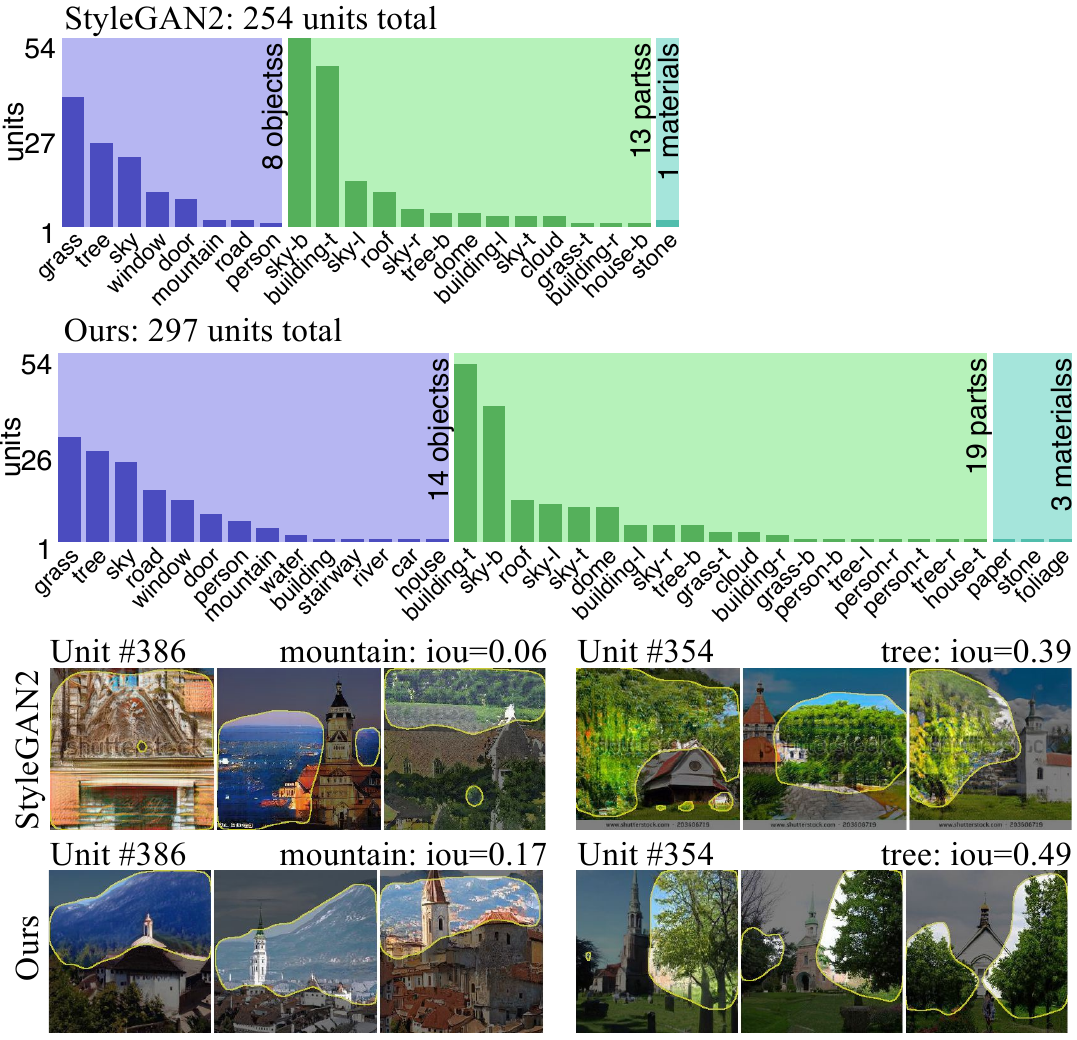}
    \caption{\textbf{GAN Dissection visualization of improved units.} We analyze StyleGAN2 and our model trained on \textsc{LSUN Church} using GAN Dissection~\cite{bau2018gan} and show here qualitative examples of units with improved IoU to a semantic category.
    The total number of detected units also increases from 254 to 297 for our model.}
    \lblfig{gan_dissection}
    \vspace{-10pt}
\end{figure}

\begin{figure}[!t]
    \centering
    \includegraphics[width=0.44\textwidth]{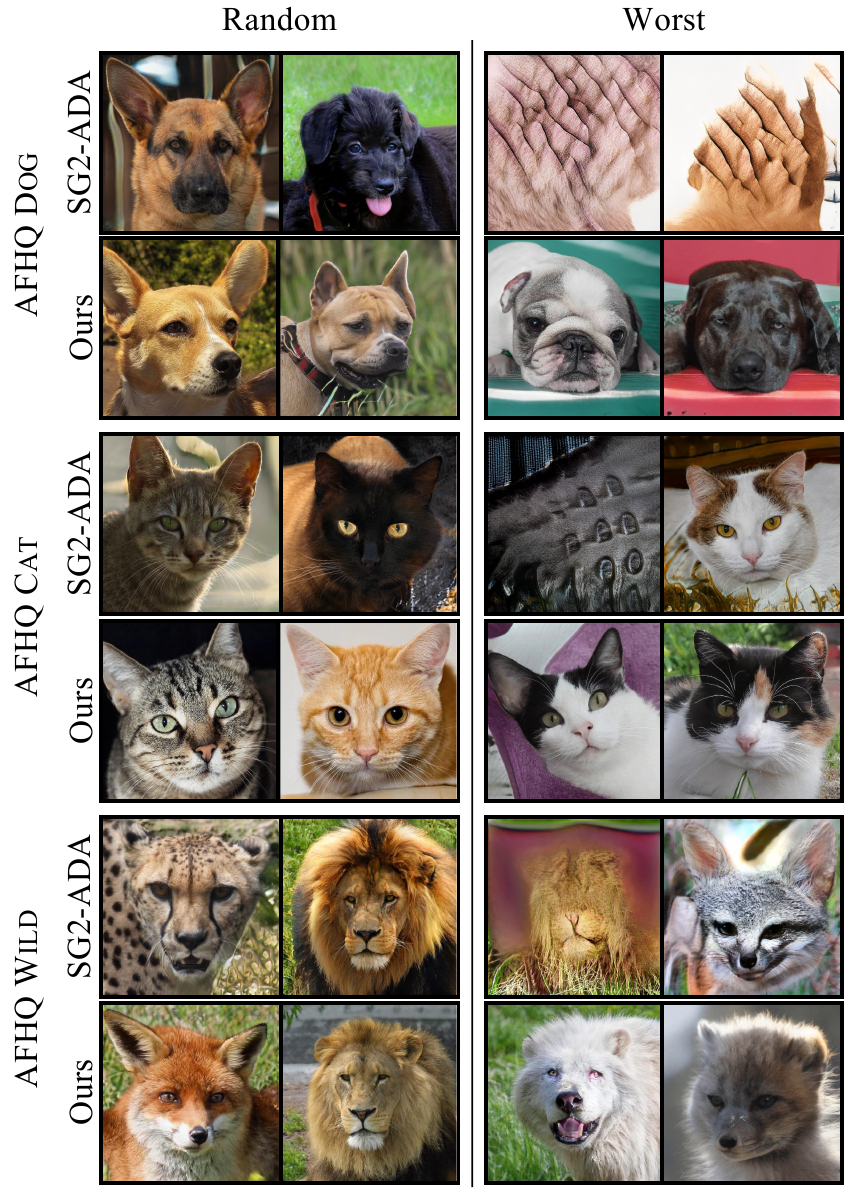}
    \caption{\textbf{Qualitative comparison of our method with StyleGAN2-ADA on AFHQ.} \textit{Left:} randomly generated samples for both methods. \textit{Right:} For both our model and StyleGAN2-ADA, we independently generate 5k samples and find the worst-case samples compared to real image distribution. We first fit a Gaussian model using the Inception~\cite{szegedy2016rethinking} feature space of real images. We then calculate the log-likelihood of each sample given this Gaussian prior and show the images with minimum log-likelihood (maximum Mahalanobis distance). We show more samples in \reffig{fig:appendix-afhq} and \reffig{fig:appendix-afhq2} in \refapp{appendix-1}.
    }\lblfig{afhq}
      \vspace{-8pt}
\end{figure}

\begin{table*}[!t]
\centering
\scalebox{0.78}{
\begin{tabular}{@{\extracolsep{4pt}}lcrrrrrrrrr@{}}
\toprule
\multirow{2}{*}{\textbf{Dataset}}
&\multirow{2}{*}{\textbf{Transfer}}
&\multicolumn{3}{c}{\textbf{StyleGAN2}}
&\multicolumn{3}{ c}{\textbf{StyleGAN2-ADA}}
&\multicolumn{3}{c}{\textbf{Ours (w/ ADA)}} \\
\cmidrule{3-5}\cmidrule{6-8}\cmidrule{9-11}
& & FID $\downarrow$ & KID $\downarrow$ & Recall $\uparrow$ & FID $\downarrow$ & KID $\downarrow$ & Recall $\uparrow$ & FID $\downarrow$& KID $\downarrow$ & Recall $\uparrow$ \\
 \midrule
\multirow{2}{*}{\textsc{AFHQ Dog}}  & \xmark & 22.35 & 10.05 & 0.20 & 7.60 & 1.29 & 0.47 & \textbf{4.73} & \textbf{0.39} & \textbf{0.60} \\ 
 & \cmark & 9.28  & 3.13 & 0.42  & 7.52& 1.22  & 0.43 & \textbf{4.81} & \textbf{0.37} & \textbf{0.61} \\ 
  \midrule
\multirow{2}{*}{\textsc{AFHQ Cat}}  & \xmark  & 5.16 & 1.72 &  0.26 & 3.29 & 0.72 & 0.41 & \textbf{2.53} & \textbf{0.47} & \textbf{0.52} \\ 
 & \cmark &  3.48 & 1.07 & 0.47 & 3.02 & \textbf{0.38} & 0.45 &  \textbf{2.69} & 0.62 & \textbf{0.50}  \\ 
  \midrule
\multirow{2}{*}{\textsc{AFHQ Wild}}  & \xmark   & 3.62 & 0.84 &  0.15 & 3.00 & 0.44 & 0.14 & \textbf{2.36}  & \textbf{0.38}  & \textbf{0.29}\\ 
 & \cmark & \textbf{2.11}  & \textbf{0.17} & 0.35 & 2.72 & 0.17 & 0.29 & 2.18 &  0.28 &  \textbf{0.38} \\ 
 \midrule
\textsc{\textsc{MetFaces}}  &  \cmark & 57.26 & 2.50 &  \textbf{0.34}  & 17.56 & 1.55  & 0.22 & \textbf{15.44} &  \textbf{1.03}  &  0.30 \\ 
\bottomrule
\end{tabular}}
\caption{\textbf{Results on AFHQ and \textsc{MetFaces}}. Our method, in general, results in lower FID and higher Recall. In transfer setup we fine-tune from a FFHQ trained model of similar resolution with $D$ updated according to FreezeD technique~\cite{mo2020FreezeD} similar to~\cite{stylegan2ada}. We select the snapshot with the best FID and show an average of three evaluations. KID is shown in $\times 10^3$ units following~\cite{stylegan2ada}.
}
\lbltbl{few-shot-afhq}
  \vspace{-10pt}
\end{table*}

\myparagraph{GAN Dissection analysis on \textsc{LSUN Church}.}
How does the generator change with the use of off-the-shelf models as discriminators? To analyze this, we use the existing technique of GAN Dissection~\cite{bau2018gan,bau2020understanding}, which calculates the correlation between convolutional feature maps of the generator and scene parts obtained through a semantic segmentation network~\cite{xiao2018unified}. Specifically, we select the convolutional layer with $32$ resolution in the generator trained with our method and StyleGAN2 on the full \textsc{LSUN Church} dataset. The total number of interpretable units~\cite{bau2018gan} increases from $254$ to $297$ by our method, suggesting that our model may learn a richer representation of semantic concepts. \reffig{gan_dissection} shows the complete statistics of detected units corresponding to each semantic category and some of the example images of improved units by our method. We observe an overall increase in the number of detected units as well as units corresponding to new semantic categories.

\myparagraph{Human preference study.} 
As suggested by ~\cite{kynkaanniemi2022role} we perform a human preference study on Amazon Mechanical Turk (AMT) to verify that our results agree with the human judgment regarding the improved sample quality. We compare StyleGAN2-ADA and our method trained on $1$k samples of \textsc{LSUN Cat}, \textsc{LSUN Church}, and \textsc{FFHQ} datasets. Since we fine-tune StyleGAN2-ADA with our method, the same latent code corresponds to similar images for the two models, as also shown in \reffig{train_vary_samples}. For randomly sampled latent codes, we show the two images generated by our method and StyleGAN2-ADA for six seconds to the test subject and ask to select the more realistic image. We perform this study for $50$ test subjects per dataset, and each subject is shown a total of $55$ images. On the \textsc{FFHQ} dataset, human preference for our method is $53.8 \% \pm 1.3$. For the \textsc{LSUN Church} dataset, our method is preferred over StyleGAN2-ADA with $60.5 \% \pm 1.7$, and for the \textsc{LSUN Cat} dataset $63.5 \% \pm 1.6$. These results correlate with the improved FID metric. Example images from our study are shown in \reffig{fig:appendix-human_study}. 

\begin{figure*}[!t]
    \centering
    \includegraphics[width=0.95\textwidth]{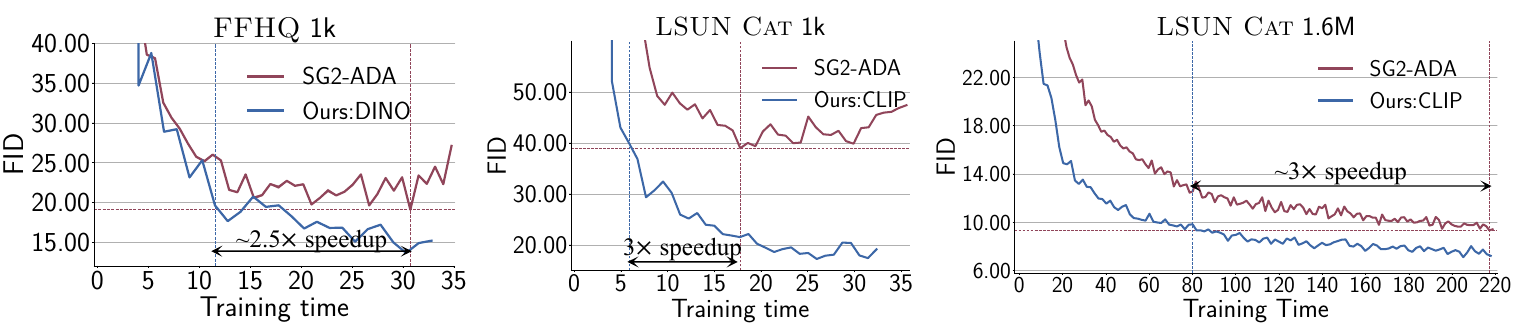} 
    \caption{\textbf{FID$\downarrow$ w.r.t. training time comparison} between StyleGAN2-ADA and our method (w/ ADA and one pretrained model) when applied from the start for \textsc{FFHQ}, \textsc{LSUN Cat} 1k, and \textsc{LSUN Cat} full-dataset setting. There is a warm-up of $0.5$M images, and then our loss is added. Our method results in similar FID at more than twice the speedup. We show training time in hours measured on one RTX 3090.}
    \lblfig{scratch_training}
     \vspace{-8pt}
\end{figure*}

\begin{table}[!t]
\centering
\setlength\tabcolsep{2pt} 
\scalebox{0.8}{
\begin{tabular}{@{\extracolsep{0pt}} l l r r @{\extracolsep{6.0pt}} r r @{\extracolsep{6.0pt}} r r@{}}
\toprule
\multicolumn{2}{c}{\multirow{2}{*}{\textbf{Method}}}
&\multicolumn{2}{c}{\textbf{Bridge}}
&\multicolumn{2}{c}{\textbf{AnimalFace Cat}}
&\multicolumn{2}{c}{\textbf{AnimalFace Dog}} \\
\cmidrule{3-4} \cmidrule{5-6} \cmidrule{7-8}
& & FID $\downarrow$ & KID$\downarrow$ & FID $\downarrow$& KID \small $\downarrow$& FID$\downarrow$ & KID$\downarrow$ \\ 
\midrule
& DiffAugment & 54.50 & 15.68 & 43.87 & 7.56 &  60.50 &  20.13 \\ 
& ADA & - & - & 38.01 & 5.61 & 52.59 &  14.32 \\ 
 \midrule
\multirow{3}{*}{\rotatebox[origin=c]{90}{\textbf{Ours}}} &  \textbf{+$1^{\text{st}}$ D} & 44.18 & 9.27 & 30.62 & 1.15  & 34.23 & 2.01  \\ 
& \textbf{+$2^{\text{nd}}$ D} & \textbf{33.89} & \textbf{2.35} & 28.01 &  0.37 &33.03  & \textbf{1.37} \\ 
& \textbf{+$3^{\text{rd}}$ D} &  34.35 &  2.96 & \textbf{27.35}  & \textbf{0.34} & \textbf{32.56}  &  1.67 \\ 
\bottomrule
\end{tabular}}
\caption{\textbf{Low-shot generation results} on 100-shot Bridge dataset~\cite{diffaug}, AnimalFace cat and dog ~\cite{si2011learning} categories. Our method significantly improves FID and KID compared to leading methods for few-shot GAN training. KID is shown in $\times10^3$ units. 
}
\lbltbl{low-shot}
 \vspace{-10pt}
\end{table}

\subsection{\textsc{AFHQ} and \textsc{MetFaces}}
To further evaluate our method on real-world limited sample datasets, we perform experiments on \textsc{MetFaces} (1336 images) and \textsc{AFHQ} dog, cat, wild categories with $\sim$ 5k images per category. We compare with StyleGAN2-ADA under two settings, (1) Fine-tuning StyleGAN2-ADA model with our loss (2) Fine-tuning from a StyleGAN2 model trained on \textsc{FFHQ} dataset of same resolution (transfer setup) using FreezeD~\cite{mo2020FreezeD}. The second setting evaluates the transfer learning capability when fine-tuned from a generator trained on a different domain. \reftbl{few-shot-afhq} shows the comparison of our method with StyleGAN2 and StyleGAN2-ADA on multiple metrics. We outperform or perform on-par compared to the existing methods in general. \reffig{afhq} shows the qualitative comparison between our method and StyleGAN2-ADA.

\subsection{Low-shot Generation}
To test our method to the limit of low-shot samples, we evaluate our method when only 100-400 samples are available. We finetune StyleGAN2 model with our method on AnimalFace cat (169 images) and dog (389 images)~\cite{si2011learning}, and 100-shot Bridge-of-Sighs~\cite{diffaug} datasets. For differentiable augmentation, we use ADA except for the 100-shot dataset where we find that the DiffAugment~\cite{diffaug} works better than ADA~\cite{stylegan2ada}, and therefore employ that. Our method leads to considerable improvement over existing methods on both FID and KID metrics as shown in \reftbl{low-shot}. We show nearest neighbour test and latent space interpolations in \reffig{fig:appendix-nn-test} and \reffig{fig:appendix-low_shot} of \refapp{appendix-1}. 
 
\subsection{Ablation Study}
\lblsec{ablation}

\myparagraph{Fine-tuning vs. training from scratch.}
In all our experiments, we fine-tuned a well-trained StyleGAN2 model (both generator and discriminator) with our additional loss. We show here that our method works similarly well when training from scratch. \reffig{scratch_training} shows the plot of FID with training time for StyleGAN2-ADA and our method with a single vision-aided discriminator on \textsc{FFHQ} and \textsc{LSUN Cat} trained with 1k samples, and \textsc{LSUN Cat} full-dataset setting. Our method results in better FID and converges more than $2\times$ faster. During training from scratch, we train with the standard adversarial loss for the first $0.5$M images and then introduce the discriminator selected by the model selection strategy. Training with three vision-aided discriminators for same number of iterations as \reftbl{few-shot-training-vary} we get similar FID of $10.60$ and $12.24$ for \textsc{FFHQ} and \textsc{LSUN Cat} 1k respectively.

\myparagraph{K-progressive vs. K-fixed model selection.}
We compare the K-progressive and K-fixed model selection strategies in this section. \reffig{progressive_vs_fixed} shows the comparison for \textsc{FFHQ} 1k and \textsc{LSUN Cat} 1k trained for the same number of iterations with two models from our model bank. We observe that training with two fixed pretrained models from the start results in a similar or slightly worse FID at the cost of extra training time compared to the progressive addition. 

\begin{figure}[t]
    \centering
    \includegraphics[width=0.48\textwidth]{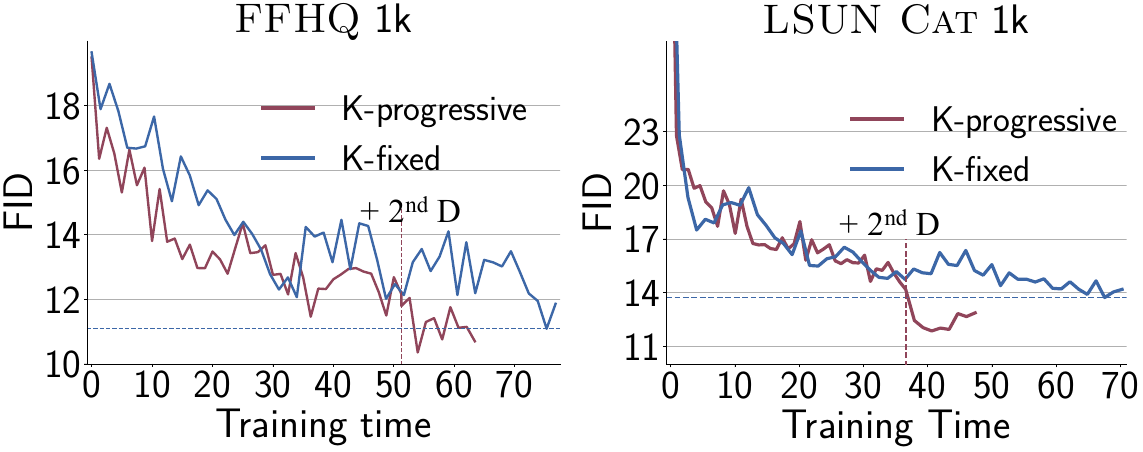}
    \caption{\textbf{K-progressive vs K-fixed comparison} on \textsc{FFHQ} and \textsc{LSUN Cat} 1k setting. Progressive addition of the next best model is computationally efficient and results in similar FID at lower run time. We show training time in hours measured on one RTX 3090.}
    \lblfig{progressive_vs_fixed}
    \vspace{-3pt}
\end{figure}

\begin{table}[t]
\centering
\setlength\tabcolsep{2pt} 
\scalebox{0.8}{
\begin{tabular}{@{\extracolsep{0pt}}l @{\extracolsep{10pt}}  
                r @{\extracolsep{6pt}} 
                r @{\extracolsep{6pt}} 
                r @{\extracolsep{12pt}} 
                r @{\extracolsep{6pt}} 
                r @{\extracolsep{6pt}} 
                r @{}
                }
\toprule
\multirow{2}{*}{\textbf{Model}} & \multicolumn{3}{c}{\textsc{FFHQ} 1k}  & \multicolumn{3}{c}{\textsc{LSUN Cat} 1k }  \\ 
\cmidrule{2-4}\cmidrule{5-7}
 \textbf{Selection} & \textbf{+$1^{\text{st}}$ D} & \textbf{+$2^{\text{nd}}$ D} & \textbf{+$3^{\text{rd}}$ D}  & \textbf{+$1^{\text{st}}$ D} &\textbf{+$2^{\text{nd}}$ D} &\textbf{+$3^{\text{rd}}$ D} \\
\midrule
Best   & 11.43 & \textbf{10.39} & 10.58  &  15.49 &  12.90 &   \textbf{12.19} \\ 
Random  & 15.48 & 12.54 & 11.92 & 19.02 & 15.12  & 14.28 \\ 
Worst & 15.48 & 15.45 & 13.88 & 19.02 & 17.53 & 17.66 \\
\bottomrule
\end{tabular}}
\caption{\textbf{FID$\downarrow$ metric for models trained with different model selection strategies in K-progressive vision-aided training.} \textit{ $1^{\text{st}}$ Row:} model selection with best linear probe accuracy. \textit{$2^{nd}$ Row:} randomly selecting from the bank of off-the-shelf models. \textit{$3^{rd}$ Row:} model selection with least linear probe accuracy.
}
\lbltbl{random_scratch_comparison}
  \vspace{-10pt}
\end{table}

\begin{figure*}[!t]
    \centering
    \includegraphics[width=0.9\textwidth]{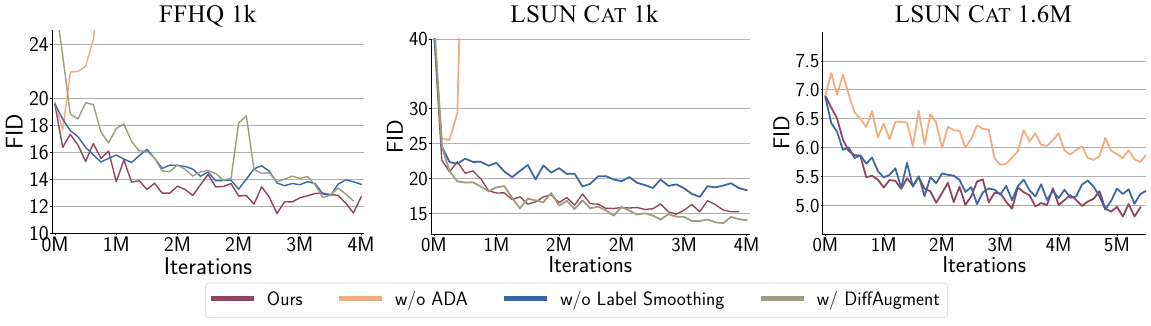}
     \caption{\textbf{Ablation of augmentation and label smoothing} on \textsc{FFHQ} and \textsc{LSUN Cat} with 1k training samples and \textsc{LSUN Cat} full-dataset setting. We show the plot of FID w.r.t training iterations when ADA~\cite{stylegan2ada} augmentation and label smoothing~\cite{improvedgan_labelsmoothing} are individually removed from our training. Without differentiable augmentation, model training quickly collapses in limited sample setting. Even for full-dataset, using differentiable augmentation for vision-aided discriminator results in better FID. Label smoothing has a reasonable effect in case of \textsc{LSUN Cat} 1k and is marginally helpful for \textsc{FFHQ} 1k. We also change the augmentation technique to DiffAugment~\cite{diffaug} for both original and vision-aided discriminator and observe that it performs comparable to ADA \cite{stylegan2ada}.}
    \lblfig{noauag_ablation}
     \vspace{-10pt}
\end{figure*}

\myparagraph{Our model selection vs. random selection.}
We showed in \reffig{linear_acc} that FID correlates with model selection ranking in vision-aided GAN training with a single pretrained model. To show the effectiveness of model selection in K-progressive strategy, we compare it with (1) random selection of models during progressive addition and (2) selection of models with least linear probe accuracy. The results are as shown in \reftbl{random_scratch_comparison}. We observe that random selection of pretrained models from the model bank already provides benefit in FID, but with our model selection, it can be improved further. Details of selected models are given in \refapp{appendix-4}.

\begin{table}[t]
\centering
\setlength\tabcolsep{6pt} 
\scalebox{0.8}{
\begin{tabular}{@{\extracolsep{0pt}}  
                l   
                r 
                r 
                r@{}}
\toprule
\textbf{Method}  & \begin{tabular}{@{}c@{}} \textsc{FFHQ} \\ 1k \end{tabular}   &  \begin{tabular}{@{}c@{}} \textsc{LSUN Cat}\\ 1k \end{tabular}  &  \begin{tabular}{@{}c@{}} \textsc{LSUN Cat}\\ 1.6M \end{tabular}  \\ 
\midrule
StyleGAN2-ADA  &  19.57 & 41.14  & 6.86 \\ 
Ours (w/ ViT (CLIP)) & \textbf{11.63} & \textbf{15.49}  & \textbf{4.61} \\ 
\midrule
Ours w/ fine-tune ViT (CLIP) & \xmark  & \xmark & \xmark \\ 
Ours w/ ViT random weights & 19.10 & 33.77  & 6.35\\ 
Ours w/ multi-discriminator & 17.59  & 37.01  & \xmark \\ 
Longer StyleGAN2-ADA  & 19.07 & 39.36 & 6.52\\  
\bottomrule
\end{tabular}}
\caption{\textbf{Additional ablation studies evaluated on FID$\downarrow$ metric}. Having two discriminators during training (frozen with random weights or trainable) or standard adversarial training for more iterations leads to only marginal benefits in FID. Thus the improvement is through an ensemble of original and vision-aided discriminators. \xmark  $\;$means FID increased to twice the baseline, and therefore, we stop the training run.
}
\lbltbl{7}
  \vspace{-10pt}
\end{table}

\myparagraph{Role of data augmentation and label smoothing.}
Here, we investigate the role of differentiable augmentation~\cite{stylegan2ada,zhao2020image,tran2020towards,diffaug} which is one of the important factors that enable the effective use of pretrained features. Label smoothing~\cite{improvedgan_labelsmoothing} further improves the training dynamics, especially in a limited sample setting. We ablate each of these component and show its contribution in \reffig{noauag_ablation} on \textsc{FFHQ} and \textsc{LSUN Cat} dataset in 1k sample setting, and \textsc{LSUN Cat} full-dataset setting. \reffig{noauag_ablation} shows that replacing ADA~\cite{stylegan2ada} augmentation strategy with DiffAugment~\cite{diffaug} in our method also performs comparably. Moreover, in the limited sample setting, without data augmentation, model collapses very early in training, and FID diverges. The role of label smoothing is more prominent in limited data setting e.g. \textsc{LSUN Cat} 1k.

\myparagraph{Additional ablation study.}
Here we further analyze the importance of our design choice. All the experiments are done on \textsc{LSUN Cat} and \textsc{FFHQ}. We compare our method with the following settings: 
(1) Fine-tuning ViT (CLIP) network as well in our vision-aided adversarial loss;
(2) Randomly initializing the feature extractor network ViT (CLIP);
(3) Training with two discriminators, where the $2^{\text{nd}}$ discriminator is of same architecture as StyleGAN2 original discriminator; (4) Training the StyleGAN2-ADA model longer for the same number of iterations as ours with standard adversarial loss. The results are as shown in \reftbl{7}. We observe that the baseline methods provide marginal improvement, whereas our method offers significant improvement over StyleGAN2-ADA, as measured by FID.

\section{Limitations and Discussion}
\lblsec{discussion}
 In this work, we propose to use available off-the-shelf models to help in the unconditional GAN training. Our method significantly improves the quality of generated images, especially in the limited-data setting. While the use of multiple pretrained models as discriminators improves the generator, it has a few limitations. First, this increases memory requirement for training. Exploring the use of efficient computer vision models~\cite{tan2019efficientnet,sandler2018mobilenetv2} will potentially make our method more accessible. Second, our model selection strategy is not ideal in the low-shot settings when only a dozen samples are available. We observe increased variance in the linear probe accuracy with sample size $\sim100$ which can lead to ineffective model selection. We plan to adopt few-shot learning~\cite{snell2017prototypical,gidaris2019boosting} methods for these settings in future. 
 
 Nonetheless, as more and more self-supervised and supervised computer vision models are readily available, they should be used to good advantage for generative modeling. This paper serves as a small step towards improving generative modeling by transferring the knowledge from large-scale representation learning.
 
\clearpage 

\myparagraph{Acknowledgment.}
We thank Muyang Li, Sheng-Yu Wang, Chonghyuk (Andrew) Song for proofreading the draft. We are also grateful to Alexei A. Efros, Sheng-Yu Wang, Taesung Park, and William Peebles for helpful comments and discussion. The work is partly supported by Adobe Inc., Kwai Inc, Sony Corporation, and Naver Corporation. 

{\small
\bibliographystyle{ieee_fullname}
\bibliography{main}
}

\appendix
\renewcommand{\thefootnote}{\arabic{footnote}}

\clearpage
\noindent{\Large\bf Appendix}
\vspace{5pt}

We show more visualizations and quantitative results to show the efficacy of our method. Namely, \refsec{appendix-1} shows qualitative comparisons between our method and leading methods for GAN training i.e. StyleGAN2-ADA~\cite{stylegan2ada} and DiffAugment~\cite{diffaug}. In \refsec{appendix-2} we show results of our vision-aided adversarial training with BigGAN architecture on CIFAR-10 and CIFAR-100 datasets. In \refsec{appendix-3}, we show more evaluation of our model on other metrics. \refsec{appendix-4} details our training hyperparameters, discriminator architectures, and selected models in each experiment. In \refsec{appendix-5}, we discuss the societal impact of our work.

\section{Qualitative Image Analysis}\lblsec{appendix-1}
In \reffig{fig:appendix-train_vary_samples}, we show more randomly generated images by StyleGAN2-ADA and our method with the same latent code for \textsc{FFHQ}, \textsc{LSUN Cat}, and \textsc{LSUN Church} 1k training sample setting similar to \reffig{train_vary_samples} of the main paper. Figure \ref{fig:appendix-train_vary_samples_diffaug} shows similar comparison between DiffAugment and our method. In many cases, our method improves the visual quality of samples compared to the baseline. 

For the human preference study conducted on the 1k sample setting, \reffig{fig:appendix-human_study} shows the sample images for the cases where users preferred our generated images or StyleGAN2-ADA generated images. \reffig{fig:appendix-random_ffhq_cat} and \reffig{fig:appendix-random_ffhq_church} show randomly generated images by our method, StyleGAN2-ADA, and DiffAugment for varying training sample settings of \textsc{FFHQ}, \textsc{LSUN Cat}, and \textsc{LSUN Church}.

For \textsc{AFHQ} and \textsc{MetFaces}, \reffig{fig:appendix-afhq} and \reffig{fig:appendix-afhq2} show the qualitative comparison between StyleGAN2-ADA and our method (similar to \reffig{afhq} in the main paper). \reffig{fig:appendix-afhq3} and \reffig{fig:appendix-afhq4} show similar comparison for \textsc{FFHQ}, \textsc{LSUN Cat}, and \textsc{LSUN Church} 1k training sample setting. We also qualitatively evaluate our low-shot trained models on nearest neighbour test from training images in \reffig{fig:appendix-nn-test}. \reffig{fig:appendix-low_shot} shows the latent interpolation of models trained by our method in the low-shot setting with $100-400$ real samples. The smooth interpolation shows that the model is probably not overfitting on the few real samples.

\begin{figure}[!t]
    \centering
    \includegraphics[width=0.4\textwidth]{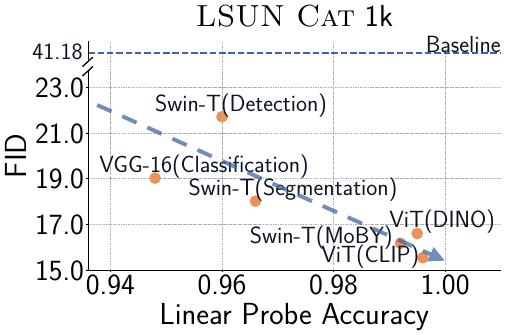}
    \caption{\textbf{Model selection using linear probing of pretrained features} on \textsc{LSUN Cat} 1k training sample setting. We show correlation of FID with the accuracy of a logistic linear model trained for real vs fake classification over the features of off-the-shelf models. Top dotted line is the FID of StyleGAN2-ADA generator used in model selection and from which we finetune with our proposed vision-aided adversarial loss. Thus, selecting models with higher linear probe accuracy in general results in better generative model with respect to FID metric.
    }\lblfig{fig:appendix-lsuncat-correlation-fid}
\end{figure}

\section{Vision-aided BigGAN}\lblsec{appendix-2}
Here, we perform experiments with BigGAN architecture~\cite{biggan} on CIFAR-10 and CIFAR-100 datasets~\cite{krizhevsky2009learning}. \reftbl{cifar} shows the comparison of vision-aided adversarial training with the current leading method DiffAugment~\cite{diffaug} in both unconditional and conditional settings, with varying training dataset sizes. We outperform DiffAugment across all settings according to the FID metric. In the case of unconditional training with BigGAN, we use self-modulation in the generator layers~\cite{morozov2020self,schonfeld2020u}. During conditional training, we use projection discriminator~\cite{miyato2018cgans} in the vision-aided discriminator as well. The training is done for the same number of iterations as DiffAugment~\cite{diffaug}.

\begin{table*}[!t]
\centering
\scalebox{0.8}{
\begin{tabular}{@{\extracolsep{4pt}}lc cccc cccc@{}}
\toprule
\multicolumn{2}{c}{\textbf{BigGAN}} & \multicolumn{4}{c}{\textbf{Conditional}}& \multicolumn{4}{c}{\textbf{Unconditional}} \\ 
& & \multicolumn{2}{c}{\textbf{CIFAR-10}} & \multicolumn{2}{c}{\textbf{CIFAR-100}} 
& \multicolumn{2}{c}{\textbf{CIFAR-10}} &\multicolumn{2}{c}{\textbf{CIFAR-100}} \\
\cmidrule{3-6} \cmidrule{7-10}
& & 100 $\%$ & 10 $\%$  & 100 $\%$  & 10 $\%$ & 100 $\%$  & 10 $\%$ & 100 $\%$  & 10 $\%$  \\
\midrule

& DiffAugment~\cite{diffaug} & 10.09  & 27.81  & 13.60 & 39.59 &  15.23 &  32.63 & 19.20 & 33.75 \\ 
& DiffAugment + CR~\cite{zhang2019consistency} & 9.68 &  22.89 & 12.65 & 30.53 & \multicolumn{4}{c}{-}  \\ 
\midrule
\multirow{3}{*}{\rotatebox[origin=c]{90}{\textbf{Ours}}} &  \textbf{+$1^{\text{st}}$ D}  & 9.93 & 17.03 & 12.46 & 23.30 & 12.41 & 21.17 & 16.08 & 25.13 \\ 
&  \textbf{+$2^{\text{nd}}$ D}  & 9.25 & 14.28 & 11.50 & 20.05 & 11.59 & 17.39& 15.05 & 20.89 \\ 
&  \textbf{+$3^{\text{rd}}$ D}  & \textbf{8.75} & \textbf{13.11} & \textbf{10.88} &\textbf{15.71} &  \textbf{11.17} &  \textbf{16.34} & \textbf{14.10} & \textbf{19.13} \\ 
\bottomrule
\end{tabular}}
\caption{Vision-aided GAN training on CIFAR-10 and CIFAR-100~\cite{krizhevsky2009learning} with the BigGAN architecture. We improve FID on both conditional and unconditional training setups. FID is calculated using \texttt{clean-fid} with $10$k generated samples and test set as the reference distribution. The three off-the-shelf models selected by model selection are CLIP, DINO, and MoBY (Swin-T), respectively, in all settings.
}
\lbltbl{cifar}
  \vspace{-10pt}
\end{table*}

\section{Evaluation}\lblsec{appendix-3}
We measure FID using \texttt{clean-fid} library~\cite{cleanfid} with $50$k generated samples and complete training dataset as the reference distribution in all our experiments similar to StyleGAN2-ADA~\cite{stylegan2ada} except in low-shot setting. For low-shot 100-400 sample setting we compute FID and KID with 5k generated samples and full available real dataset as the reference following DiffAugment~\cite{diffaug}. In addition to the FID metric reported in the main paper, we report in \reftbl{few-shot-training-vary-pr} and \reftbl{full-dataset-pr}, precision and recall metrics~\cite{improvedprecisionrecall} for \textsc{FFHQ} and \textsc{LSUN} experiments with varying training sample size and full-dataset. We observe that our method improves the recall metric in all cases and has similar or better precision, particularly in the limited sample settings. Recent studies~\cite{kynkaanniemi2022role,morozov2020self} suggest reporting FID metrics in different feature spaces, in order to avoid ``attacking'' the evaluation metric. As such, we also report FID, but using the feature space of a self-supervised model trained on ImageNet via SwAV~\cite{morozov2020self}, in \reftbl{swav_eval} to \reftbl{swav_eval4}, and observe consistent improvements. Moreover, as shown in \reftbl{few-shot-training-vary} and \reftbl{multiscale}, when using CLIP (which is not trained on ImageNet) or using DINO (a self-supervised method that does not require ImageNet labels) in our vision-aided training, we also improve FID scores. \reftbl{few-shot-afhq-all} shows the results on progressive addition of vision-aided discriminator on \textsc{MetFaces} and AFHQ.

\begin{table}[!t]
\centering
\scalebox{.9}{
\begin{tabular}{lcc}
\toprule
\textbf{Discriminator Architecture} & \multicolumn{2}{c}{\textbf{Dataset}} \\
& \textsc{LSUN Cat} 1k & \textsc{FFHQ} 1k \\
\midrule
Single-scale (MLP) & 17.21 & 13.31 \\
Multi-scale & \textbf{15.49} & \textbf{11.63} \\
\bottomrule
\end{tabular}}
\caption{\textbf{FID$\downarrow$ on Single-scale vs Multi-scale discriminator head for ViT (CLIP)}. We observe slightly better FID with a multi-scale discriminator head compared to a 2-layer MLP head on final classification token feature, without any significant increase in training time. Therefore, we select the multi-scale discriminator head for all our experiments on both CLIP and DINO with ViT-B architecture.
}
\lbltbl{multiscale}
\end{table}

\begin{table*}[!t]
\centering
\setlength{\tabcolsep}{5pt}
\scalebox{.9}{
\begin{tabular}{ @{\extracolsep{0pt}} l r  rr rr  rr  rr rr rr @{}}
\toprule
\multicolumn{2}{l}{\multirow{2}{*}{\textbf{Dataset}}}
& \multicolumn{2}{c}{\multirow{2}{*}{\textbf{StyleGAN2}}}
& \multicolumn{2}{c}{\multirow{2}{*}{\textbf{DiffAugment}}}
& \multicolumn{2}{c}{\multirow{2}{*}{\textbf{ADA}}}
&\multicolumn{6}{c}{\textbf{Ours (w/ ADA)}}  \\
\cmidrule{9-14}
& & & & & & & & \multicolumn{2}{c}{\textbf{+$1^{\text{st}}$ D}} & \multicolumn{2}{c}{\textbf{+$2^{\text{nd}}$ D}} & \multicolumn{2}{c}{\textbf{+$3^{\text{rd}}$ D}} \\
\midrule
& & P $\uparrow$ & R $\uparrow$& P $\uparrow$ & R $\uparrow$&P $\uparrow$ & R $\uparrow$&P $\uparrow$ & R $\uparrow$&P $\uparrow$ & R $\uparrow$&P $\uparrow$ & R $\uparrow$ \\
\midrule
\multirow{3}{*}{\rotatebox{90}{\textbf{FFHQ}}} 
& 1k   & 0.580 & 0.000 & 0.681  & 0.034 & 0.675 & 0.088 & 0.719 & 0.139 & \textbf{0.740} & 0.139 & 0.694 & \textbf{0.173} \\ 
& 2k   & 0.586 &  0.025  & 0.709 & 0.099 & 0.676 & 0.137 & 0.701 & 0.242 & 0.719 &  0.251 & \textbf{0.719} & \textbf{0.251}   \\ 
& 10k  & 0.669 &  0.191  &  0.704 &  0.256  & 0.700 & 0.255 & 0.683 & 0.334 & 0.687 & 0.342 & \textbf{0.697} &  \textbf{0.351} \\ 
\midrule
\multirow{3}{*}{\rotatebox{90}{\begin{tabular}{@{}c@{}} \textsc{\textbf{LSUN}}\\ \textsc{\textbf{Cat}} \end{tabular} }} 
& 1k  & 0.290 & 0.000 & 0.539 & 0.015 &  0.468 & 0.012 & \textbf{0.653} &  0.033 & 0.627 & 0.053 & 0.624 &  \textbf{0.058} \\ 
& 2k   &  0.527 &  0.001  & 0.607  & 0.032 &  0.617 &  0.066 & \textbf{0.667} &  0.084 & .645 &  0.105 & 0.652 & \textbf{0.120}  \\ 
& 10k  & 0.632 & 0.099 & 0.628 & 0.188 &  0.598 &  0.111 & \textbf{0.639} & 0.152 & 0.615 & 0.181 & 0.599 &  \textbf{0.203} \\ 
\midrule
\multirow[c]{3}{*}{\rotatebox{90}{\begin{tabular}{@{}c@{}} \textbf{LSUN}\\ \textbf{Church} \end{tabular} }} 
& 1k   & - & - & 0.593 & 0.015 & 0.554 & 0.038 & \textbf{0.652} & 0.047 & 0.609 &  \textbf{0.065} & 0.645 & 0.063 \\ 
& 2k   &  & & 0.613 & 0.042 & 0.604 &  0.075 & 0.637 & 0.100 & 0.626&  0.107 & \textbf{0.649} & \textbf{0.114}  \\ 
& 10k  & - & - & 0.567 & 0.256 & 0.617 & 0.108 & \textbf{0.662} & 0.132 &  0.645 & 0.112 & 0.643 & \textbf{0.133}
\\ 
\bottomrule
\end{tabular}}
\caption{\textbf{Precision (P) and Recall (R) metrics for experiments on FFHQ and \textsc{LSUN} datasets} with varying training samples from 1k to 10k. Complete training dataset is used as the reference distribution for calculating the above metrics and average of 3 evaluation runs is reported. Our method results in higher recall and precision in all settings. In addition, we observe that as we add vision-aided discriminators, recall increases at the cost of slight decrease in precision.
}
\lbltbl{few-shot-training-vary-pr}
\end{table*}

\begin{table*}[!t]
\centering
\scalebox{0.8}{
\begin{tabular}{@{\extracolsep{4pt}}ll ccc @{\hspace{2\tabcolsep}} ccc ccc ccc@{}}
\toprule
\multirow{2}{*}{\textbf{Dataset}}
&\multirow{2}{*}{\textbf{Resolution}}
&\multicolumn{3}{c}{\textbf{StyleGAN2 (F)}} 
&\multicolumn{9}{c}{\textbf{Ours (w/ ADA)}} \\
& & & & & \multicolumn{3}{c}{\textbf{+$1^{\text{st}}$ D}} & \multicolumn{3}{c}{\textbf{+$2^{\text{nd}}$ D}} & \multicolumn{3}{c}{\textbf{+$3^{\text{rd}}$ D}} \\
\cmidrule{3-5} \cmidrule{6-8} \cmidrule{9-11} \cmidrule{12-14}
& & FID $\downarrow$ & P $\uparrow$ & R $\uparrow$ & FID $\downarrow$ & P$\uparrow$ & R $\uparrow$ & FID $\downarrow$ & P$\uparrow$ & R $\uparrow$ & FID $\downarrow$ & P$\uparrow$ & R $\uparrow$  \\
\midrule

\textsc{FFHQ} & 1024 $\times$ 1024  & \textbf{2.98} & \textbf{0.684} & 0.495 & 3.11 & 0.656 & \textbf{0.519} & 3.01 & 0.678 & 0.499 & 3.09 & 0.665 & 0.507 \\ 
\textsc{LSUN Cat} & 256 $\times$ 256 & 6.86 & 0.606 & 0.318 & 4.61  &\textbf{0.626} & 0.341 & 4.19  & 0.608 & 0.355 & \textbf{3.98} & 0.598 & \textbf{0.381} \\ 
\textsc{LSUN Church} & 256 $\times$ 256 & 4.28 & 0.594 & 0.391 & 2.05 & 0.596 & 0.447 & 1.81 & 0.603  & 0.451 & \textbf{1.72}   &  \textbf{0.612} &  \textbf{0.451} \\ 
\textsc{LSUN Horse} & 256 $\times$ 256 & 4.09 & 0.609 & 0.357 &  2.79 & \textbf{0.636} & 0.369 &  2.38 & 0.614  & 0.406 & \textbf{2.11}   &  0.611 & \textbf{0.416} \\ 
\bottomrule
\end{tabular}}
\caption{\textbf{FID}, \textbf{Preicison (P), and Recall (R) metrics on full-dataset setting}. Our method results in improved recall for most cases and has similar precision compared to StyleGAN2. A higher recall is usually preferred as with truncation precision can be recovered~\cite{stylegan2}.
}
\lbltbl{full-dataset-pr}
\end{table*}

\begin{table}[!t]
\centering
\setlength{\tabcolsep}{5pt}
\scalebox{0.73}{
\begin{tabular}{ @{\extracolsep{0pt}} l@{\hspace{0.2\tabcolsep}}r @{\hspace{0.2\tabcolsep}}c@{\hspace{0.98\tabcolsep}} c  c  c c  @{}}
\toprule
\multicolumn{2}{l}{\multirow{1}{*}{\textbf{Dataset}}}
& \multirow{1}{*}{\textbf{StyleGAN2}}
& \multirow{1}{*}{\textbf{DiffAugment}}
& \multirow{1}{*}{\textbf{ADA}}
&\multicolumn{1}{c}{\begin{tabular}{@{}c@{}} \textbf{Ours} \\ (w/ ADA)\end{tabular} } 
&\multicolumn{1}{c}{\begin{tabular}{@{}c@{}} \textbf{Ours} \\ (w/ DiffAugment)\end{tabular}  } \\
\midrule
\multirow{3}{*}{\rotatebox{90}{\textbf{FFHQ}}} 
& 1k   & 14.42  & 6.31 & 4.15 & \textbf{1.18} & 1.95 \\ 
& 2k   & 7.72 & 3.53 & 3.32 & \textbf{0.91} & 1.27 \\ 
& 10k   & 3.85 & 1.95 & 1.59 & \textbf{0.61} & 0.73  \\ 
\midrule
\multirow{3}{*}{\rotatebox{90}{\begin{tabular}{@{}c@{}} \textbf{LSUN}\\ \textbf{\textsc{Cat}} \end{tabular} }} 
& 1k    & 22.71 & 10.74 & 10.33 & 3.27 & \textbf{2.82}  \\ 
& 2k     & 14.97 & 7.90 & 6.23 & 3.10 & \textbf{2.57} \\ 
& 10k   & 6.92 & 4.97 & 4.50 & \textbf{1.78} & 1.96  \\ 
\midrule
\multirow{3}{*}{\rotatebox[]{90}{ \begin{tabular}{@{}c@{}} \textbf{LSUN}\\ \textbf{\textsc{\small{Church} }} \end{tabular} }} 
& 1k   & - & 7.66 & 6.49 & \textbf{2.50} & 2.87 \\ 
& 2k    & - & 6.28 & 4.54 & \textbf{1.49} & 1.57 \\ 
& 10k    & - & 3.43 & 3.25 & 1.24 & \textbf{0.98} \\  \\
\bottomrule
\end{tabular}}

\caption{\textbf{SwAV-FID~\cite{morozov2020self} of models trained on FFHQ, \textsc{LSUN} datasets with varying training samples}. FID$\downarrow$ is measured in SwAV ResNet-50 feature space with complete dataset as reference distribution. We select the best snapshot according to training set FID, and report mean of 3 FID evaluations. In Ours (w/ ADA) we finetune the pretrained StyleGAN2-ADA model, and in Ours (w/ DiffAugment) we finetune the model trained with DiffAgument while using the corresponding policy for augmentation.
}
\lbltbl{swav_eval}
\end{table}

\begin{table}[!t]
\centering
\scalebox{0.8}{
\begin{tabular}{@{\extracolsep{4pt}}lccc@{}}
\toprule
\multirow{1}{*}{\textbf{Dataset}}
&\multirow{1}{*}{\textbf{Transfer}}
&\multicolumn{1}{ c}{\textbf{StyleGAN2-ADA}}
&\multicolumn{1}{c}{\begin{tabular}{@{}c@{}} \textbf{Ours} \\ (w/ ADA)\end{tabular} } \\
\midrule
\multirow{2}{*}{\textsc{AFHQ Dog}}  & \xmark & 2.02 & \textbf{1.04} \\ 
 & \cmark & 1.89  & \textbf{1.03}  \\ 
  \midrule
\multirow{2}{*}{\textsc{AFHQ Cat}}  & \xmark  & 1.17 & \textbf{0.62}  \\ 
 & \cmark &  0.98 & \textbf{0.70} \\ 
  \midrule
\multirow{2}{*}{\textsc{AFHQ Wild}}  & \xmark  &  1.89 & \textbf{1.10} \\ 
 & \cmark & 1.23  & \textbf{0.97} \\ 
 \midrule
\textsc{\textsc{MetFaces}}  &  \cmark  & 2.14 & \textbf{1.72} \\ 
\bottomrule
\end{tabular}
}
\caption{\textbf{SwAV-FID~\cite{morozov2020self} of models trained on AFHQ categories and \textsc{MetFaces}}. FID$\downarrow$ is measured in SwAV ResNet-50 feature space with complete dataset as reference distribution. We select the best snapshot according to training set FID, and report mean of 3 FID evaluations.  In transfer setup we fine-tune from a FFHQ trained model of similar resolution with $D$ updated according to FreezeD technique~\cite{mo2020FreezeD} similar to~\cite{stylegan2ada}. 
}
\lbltbl{swav_eval2}
\end{table}

\begin{table}[!t]
\centering
\scalebox{0.8}{
\begin{tabular}{@{\extracolsep{4pt}}lcc@{}}
\toprule
\multirow{1}{*}{\textbf{Dataset}}
&\multicolumn{1}{c}{\textbf{StyleGAN2 (F)}} 
&\multicolumn{1}{c}{\textbf{Ours (w/ ADA)}}  \\
\midrule
\textsc{FFHQ-1024}  & 0.57  & \textbf{0.38}  \\ 
\textsc{LSUN Cat-256}  & 2.65 & \textbf{1.03}  \\ 
\textsc{LSUN Church-256} & 1.81 & \textbf{0.58} \\ 
\textsc{LSUN Horse-256} & 1.65 & \textbf{0.71}  \\ 
\bottomrule
\end{tabular}
}
\caption{\textbf{SwAV-FID~\cite{morozov2020self} of models trained on full dataset of FFHQ and \textsc{LSUN} categories}. FID$\downarrow$ is measured in SwAV ResNet-50 feature space with complete dataset as reference distribution. We select the best snapshot according to training set FID, and report mean of 3 FID evaluations.
}
\lbltbl{swav_eval3}
\end{table}

\begin{table}[!t]
\centering
\scalebox{0.7}{
\begin{tabular}{@{\extracolsep{0pt}} l r r @{\extracolsep{6.0pt}} r r @{\extracolsep{6.0pt}} r r@{}}
\toprule
\multicolumn{1}{c}{\multirow{2}{*}{\textbf{Method}}}
&\multicolumn{2}{c}{\textbf{Bridge}}
&\multicolumn{2}{c}{\textbf{AnimalFace Cat}}
&\multicolumn{2}{c}{\textbf{AnimalFace Dog}} \\
\cmidrule{2-3} \cmidrule{4-5} \cmidrule{6-7}
 & FID \small $\downarrow$ & KID \small $\downarrow$ & FID \small $\downarrow$& KID \small $\downarrow$& FID \small $\downarrow$ & KID \small $\downarrow$ \\ 
\midrule
DiffAugment & 8.83  & 4.96 & 12.14 & 5.62 & 18.48   &  7.56 \\ 
ADA & - & - & 10.87  &4.55  &  15.60 & 5.41  \\ 
Ours & \textbf{3.46} & \textbf{0.22} & \textbf{5.18} &  \textbf{0.29} & \textbf{6.47} & \textbf{0.35}  \\
\bottomrule
\end{tabular}

}
\caption{\textbf{SwAV-FID~\cite{morozov2020self} and KID of models trained on low-shot datasets}. FID and KID are measured in SwAV ResNet-50 feature space with complete dataset as reference distribution and 5k generated images. We select the best snapshot according to training set FID, and report mean of 3 FID and KID evaluations. KID is shown in $\times10^3$ units.
}
\lbltbl{swav_eval4}
\end{table}

\begin{table*}[!t]
\centering
\scalebox{0.8}{
\begin{tabular}{@{\extracolsep{4pt}}lc rrr rrr rrr@{}}
\toprule
\multirow{3}{*}{\textbf{Dataset}}
&\multirow{3}{*}{\textbf{Transfer}}
&\multicolumn{9}{c}{\textbf{Ours (w/ ADA)}} \\
& & \multicolumn{3}{c}{\textbf{+$1^{\text{st}}$ D}} & \multicolumn{3}{c}{\textbf{+$2^{\text{nd}}$ D}} & \multicolumn{3}{c}{\textbf{+$3^{\text{rd}}$ D}} \\
\cmidrule{3-5}\cmidrule{6-8}\cmidrule{9-11}
& & FID $\downarrow$& KID $\downarrow$ & Recall $\uparrow$ & FID $\downarrow$& KID $\downarrow$ & Recall $\uparrow$ & FID $\downarrow$& KID $\downarrow$ & Recall $\uparrow$ \\
 \midrule
\multirow{2}{*}{\textsc{AFHQ Dog}}  & \xmark &  5.67 & 0.61  & 0.54 & 4.82 &  \textbf{0.33} & 0.58 & \textbf{4.73} & 0.39 & \textbf{0.60} \\ 
 & \cmark &  5.86 & 0.70 & 0.54 & 5.08 & 0.41 &  0.59 & \textbf{4.81} & \textbf{0.37} & \textbf{0.61} \\ 
  \midrule
\multirow{2}{*}{\textsc{AFHQ Cat}}  & \xmark  & 2.95 & 0.57  & 0.46  & 2.70 & 0.61 & 0.49 & \textbf{2.53} & \textbf{0.47} & \textbf{0.52} \\ 
 & \cmark &  2.93 & 0.82 & 0.48 & 2.93 & 0.94 & 0.50 &  \textbf{2.69} & \textbf{0.62} & \textbf{0.50}  \\ 
  \midrule
\multirow{2}{*}{\textsc{AFHQ Wild}}  & \xmark   & 2.82 & 0.38   & 0.18  & 2.51 & 0.41 & 0.24 & \textbf{2.36}  & \textbf{0.38}  & \textbf{0.29} \\ 
 & \cmark & 2.26  &  0.34 & 0.35 &  2.18 & 0.28   & 0.34  & \textbf{2.18} &  \textbf{0.28} &  \textbf{0.38} \\ 
 \midrule
\textsc{\textsc{MetFaces}}  &  \cmark & 17.10 & 2.18 & 0.30  & 15.82 &  1.37 & 0.29  & \textbf{15.44} &  \textbf{1.03}  &  \textbf{0.30} \\ 
\bottomrule
\end{tabular}}
\caption{\textbf{Results on AFHQ and \textsc{MetFaces}} with progressive addition of vision-aided discriminators. In transfer setup we fine-tune from a FFHQ trained model of similar resolution with $D$ updated according to FreezeD technique~\cite{mo2020FreezeD} similar to~\cite{stylegan2ada}. We select the snapshot with the best FID and show an average of three evaluations. KID is shown in $\times 10^3$ units following~\cite{stylegan2ada}.
}
\lbltbl{few-shot-afhq-all}
\end{table*}

\begin{table*}[!t]
\centering
\scalebox{.82}{
\begin{tabular}{l  l   c c c }
\toprule
\textbf{Vision task} & \textbf{Network} & \textbf{Params} & \textbf{Extracted feature size} & \textbf{$D_i$ Architecture} \\
\midrule
ImageNet~\cite{deng2009imagenet} classifier &  VGG-16~\cite{zhang2019shiftinvar} & 138M & $512\times7\times7$ &  
\multirow{6}{*}{ $\begin{Bmatrix} \text{2}\times \text{avg. downsampling} \\ 
                \text{Conv3x3: ch} \rightarrow \text{256} \\
                \text{LeakyReLU(0.2)}\\
                \text{Linear: 256}\times \text{h} \times \text{w} \rightarrow \text{256} \\
                \text{LeakyReLU(0.2)}\\
                \text{Linear: 256}\rightarrow \text{1} \end{Bmatrix} $ } \\
MoBY~\cite{xie2021self} &  tiny Swin-T & 29M& $768\times7\times7$  &  \\ 
Face parsing~\cite{lee2020maskgan} & U-Net & 1.9M & $256\times8\times8$  &  \\
Face normals~\cite{facenormal} & U-Net + ResNet & 35M &  $512\times8\times8$ & \\ 
Segmentation~\cite{liu2021swin} & tiny Swin-T & 29M&  $768\times8\times8$ &  \\
Object detection~\cite{liu2021swin} & tiny Swin-T & 29M & $768\times8\times8$ &   \\ 
\midrule
CLIP~\cite{radford2021learning}  & ViT-B32 & 86M &
$\begin{matrix} 768\times7\times7 \\ 
768\times7\times7 \\ 
512  \end{matrix}$   & $2\times$ $\begin{Bmatrix}
                \text{Conv3x3: ch} \rightarrow \text{256} \\
                \text{LeakyReLU(0.2)}\\
                \text{2}\times \text{avg. downsample} \\ 
                \text{Conv3x3: 256} \rightarrow \text{1}  \end{Bmatrix} $, $\begin{Bmatrix}
                \text{Linear: 512} \rightarrow \text{256} \\
                \text{LeakyReLU(0.2)}\\
                \text{Linear: 256} \rightarrow \text{1} \end{Bmatrix} $ 
                  \\ \\
DINO~\cite{caron2021emerging} &  ViT-B16 &  85M & 
$\begin{matrix} 768\times14\times14 \\ 768\times14\times14 \\ 768  \end{matrix}$  &
$2\times$ $\begin{Bmatrix}
                \text{2}\times \text{avg. downsample} \\
                \text{Conv3x3: ch} \rightarrow \text{128} \\
                \text{LeakyReLU(0.2)}\\
                \text{2}\times \text{avg. downsample} \\ 
                \text{Conv3x3: 128} \rightarrow \text{1}  \end{Bmatrix} $, 
$\begin{Bmatrix}
                \text{Linear: 768} \rightarrow \text{128} \\
                \text{LeakyReLU(0.2)}\\
                \text{Linear: 128} \rightarrow \text{1} \end{Bmatrix} $
                 \\

\bottomrule
\end{tabular}}
\vspace{-5pt}
\caption{\textbf{Off-the-shelf Model Bank.} We select state-of-the-art feature extractors and task specific networks to use as an ensemble of off-the-shelf discriminators during GAN training. We keep the discriminator head architecture small and fairly similar across different models. In the multi-scale architecture of CLIP and DINO, we extract the spatial features from 4 and 8 layers and final classification token feature. In case of conditional training for CIFAR-10 and CIFAR-100 we use an additional embedding layer for number of classes and employ projection discriminator~\cite{miyato2018cgans}.}
\lbltbl{cv_models}
\vspace{-6pt}
\end{table*}

\section{Training and Hyperparameter details}\lblsec{appendix-4}
\paragraph{Off-the-shelf models and discriminator head architecture.}
We provide network details of off-the-shelf models we used in our experiments in \reftbl{cv_models}. For extracting features, we resize both real and fake images to the resolution that the pretrained network was trained on. For the trainable discriminator head, we use a \texttt{Conv-LeakyReLU-Linear-LeakyReLU-Linear} architecture over the spatial features after $2\times$ downsampling for all pretrained models except CLIP and DINO. In the case of CLIP and DINO with ViT-B architecture, we observed that a multi-scale architecture leads to marginally better results (as shown in \reftbl{multiscale}). We extract the spatial features at $4$ and $8$ layers and the final classifier token feature. For each spatial feature, we use a \texttt{Conv-LeakyReLU-Conv} with downsampling to predict a $3\times3$ real vs fake logits similar to PatchGAN~\cite{isola2017image}, and the loss is averaged over the $3\times3$ spatial grid. On the classifier token, we use \texttt{Linear-LeakyReLU-Linear} discriminator head for a global real vs. fake prediction. The final loss is the sum of losses at the three scales. Extracted feature size for each model and exact architecture of the trainable discriminator head is detailed in \reftbl{cv_models}. 

\paragraph{Linear accuracy analysis of all experiments}
\reffig{fig:appendix-linear_acc_all1} - \reffig{fig:appendix-linear_acc_all5} show the linear probe accuracy of the pretrained models and the selected model based on that. We calculate linear probe accuracy on the average of 3 runs (variance is always less than $1.\%$ except in $100$-$400$ low-sample setting where it increases to $\sim 5-8\%$ ). For the limited sample setting, we use the complete set of real training samples and the same amount of generated samples. For the full-dataset setting, we randomly sample a subset of 10k real and generated samples during linear probe accuracy calculation. We observe diminished variance in the linear classifier validation accuracy with the increase in sample size. The computational cost of calculating linear probe accuracy for a model varies with the sample size and dataset resolution but is always in the order of $5-10$ minutes, including the time for fake image generation for training linear classifier as measured on one RTX 3090. 

\paragraph{Details of our model selection vs random selection experiment in \refsec{ablation}}
In \textsc{FFHQ} 1k, our method selects DINO, CLIP, and Swin-T (MoBY) during training. Random selection consists of VGG-16, Swin-T (MoBY), and U-Net (Face Parsing) networks and worst selection consists of VGG-16, U-Net (Face Parsing), and U-Net (Face Normals) networks. For \textsc{LSUN Cat} 1k setting, CLIP, DINO, and Swin-T (Segmentation) networks are selected by our method during training. In random selection VGG-16, CLIP, and Swin-T (Segmentation) networks are selected and worst selection consists of VGG-16, Swin-T (Segmentation), and Swin-T (Detection) networks.

\vspace{-8pt}
\paragraph{Training hyperparameters and memory requirement.}
We keep similar architecture and training hyperparameters as StyleGAN2-ADA~\cite{stylegan2ada}. For experiments on \textsc{FFHQ}, \textsc{LSUN Cat}, and \textsc{LSUN Church} with varying sample size, number of feature maps at shallow layers is halved~\cite{stylegan2ada}. For $256$ resolution datasets, the weight of $R_1$ regularization ($\gamma$) in the original discriminator is $1$, learning rate is $0.002$ and path length regularization is $2$. For datasets with $512$ resolution, $\gamma$ is $0.5$ and learning rate is $0.0025$. When using ADA in our vision-aided adversarial loss, we employ cutout + bgc~\cite{stylegan2ada} policy for augmentation, and if the linear probe accuracy of the selected model is above $90\%$ one-sided label smoothing~\cite{improvedgan_labelsmoothing} is used as a regularization. The ADA target value for the original discriminator in StyleGAN2-ADA is kept the same at $0.6$. For vision-aided discriminators, we use $0.3$ as the target probability for ADA in all limited data experiments. In case of finetuning from the StyleGAN2 model in the full-dataset setting on \textsc{FFHQ} and \textsc{LSUN} categories, the original discriminator has non-augmented real and fake images as input and ADA target for additional discriminators is $0.1$. In case of training with DiffAugment, we always use one-sided label smoothing and all three augmentations color, translation, and cutout. All experiments are done with a batch size of $16$ (mini-batch std $4$) on a single RTX 3090 GPU for $256$ resolution, and $4$ GPUs for $512$, $1024$ resolution datasets. In the case of \textsc{LSUN Horse}, we fine-tuned StyleGAN2 (config F) model with a batch size of $64$ and use $\gamma$ value of $100$ following \cite{stylegan2}.

In all experiments, we train with an ensemble of three vision-aided discriminators. The number of training iterations after which we add the second model is $1$ million ($1$M) in low-shot generation settings, $4$M for 1k training sample setting, and $8$M for the rest. With the second and third pretrained model, we train for $1$M training iterations on $<1$k training sample setting and $2$M otherwise. The GPU memory requirement of our method is maximum when using VGG-16 model at $\sim2.5$GB. Next, CLIP and DINO model with ViT-B architecture have a memory requirement of $\sim2$GB. Pretrained models based on tiny Swin-T architecture and face normals and parsing model lead to $<1$GB of overhead in GPU memory during training. Compared to training the StyleGAN2 (config F) model architecture on $256$ resolution images which required $\sim 10.5$GB of GPU memory on a single RTX 3090, the overhead in memory with a single pretrained model is $10-25\%$ approximately. The maximum overhead in memory is when CLIP, DINO, and Swin-T based models are selected by model selection strategy. This results in $\sim 4.5$GB of additional memory requirement as measured on one RTX 3090 while training with our default batch-size of $16$ on $256$ resolution dataset. In the future, we hope to explore the use of efficient~\cite{tan2019efficientnet, sandler2018mobilenetv2} computer vision models to reduce the increased memory requirement of our method.

\section{Societal Impact}\lblsec{appendix-5}
Our proposed method is towards improving the image quality of GANs, specifically in the limited sample setting. This can help users in novel content creation where usually only few relevant samples are available for inspiration. Also, the faster convergence of our method when used in training from scratch makes it accessible to a broad set of people as the model can be trained at a lower computational cost. This can lead to negative societal impact as well through the creation of fake data and disinformation. One of the possible solutions to mitigate this can be to ensure reliable detection of fake generated data~\cite{wang2020cnn,chai2020makes}.

\section{Change log}
\myparagraph{v1:} Original draft.

\begin{figure*}[!t]
    \centering
    \includegraphics[width=\textwidth]{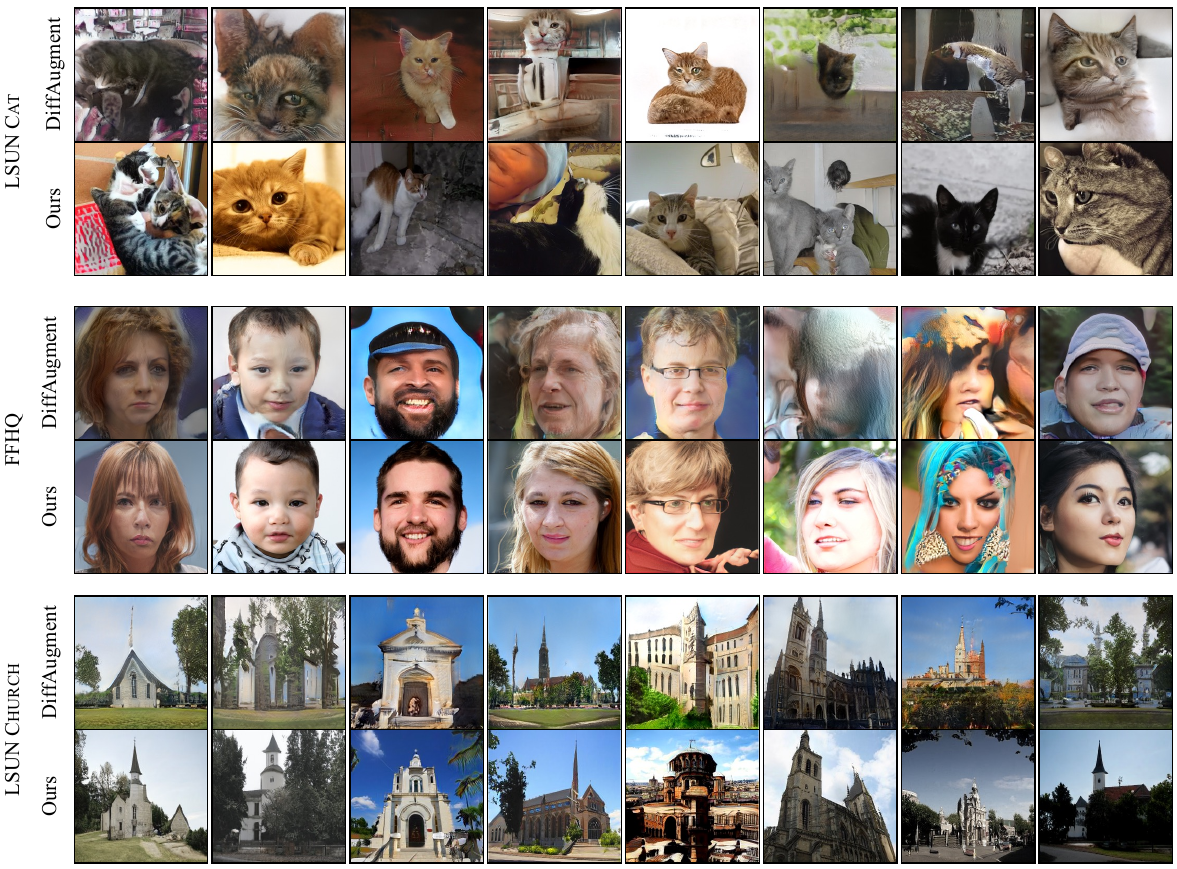}
    \captionof{figure}{\textbf{\textsc{LSUN Cat}, \textsc{FFHQ}, and \textsc{LSUN Church} paired sample comparison in 1k training dataset setting with DiffAugment}. For each dataset, the top row shows random samples of the baseline DiffAugment model with StyleGAN2 architecture, and the bottom row shows the samples by our method for the same latent code. We fine-tune StyleGAN2-DiffAugment model with our vision-aided adversarial loss. On average we observe improved image quality with our method for the same latent code. }
    \label{fig:appendix-train_vary_samples_diffaug}
\end{figure*}

\myparagraph{v2:} We included additional visualization and revised text in experiments and appendix section. Specifically, we added Figure \ref{fig:appendix-train_vary_samples_diffaug} to show qualitative comparison between our method and DiffAugment. We also added \reftbl{few-shot-afhq-all} in Appendix to show intermediate results with progressive addition of pretrained models for \textsc{MetFaces} and AFHQ categories. \reffig{fig:appendix-linear_acc_all1} - \reffig{fig:appendix-linear_acc_all5} in Appendix is updated to include linear probe accuracy plots corresponding to experiments with DiffAugment. We also updated relevant citations.

\myparagraph{v3:} We included additional results on CIFAR datasets with BigGAN in \reftbl{cifar}. We also added the FID evaluation using SwAV model in \reftbl{swav_eval} to \reftbl{swav_eval4} and nearest neighbour test for low-shot models in \reffig{fig:appendix-nn-test}.

\begin{figure*}[!t]
    \centering
    \includegraphics[width=\textwidth]{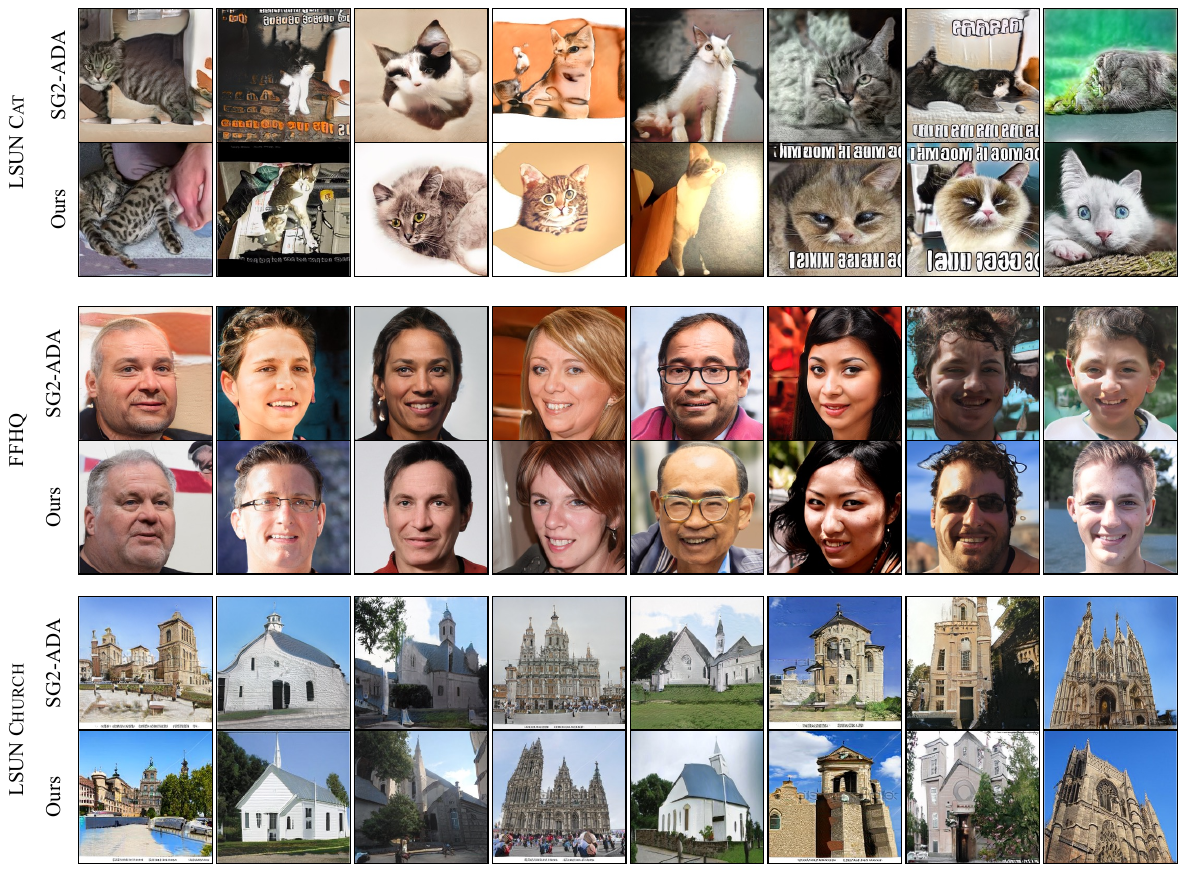}
    \captionof{figure}{\textbf{\textsc{LSUN Cat}, \textsc{FFHQ}, and \textsc{LSUN Church} paired sample comparison in 1k training dataset setting with ADA}. For each dataset, the top row shows random samples of the baseline StyleGAN2-ADA, and the bottom row shows the samples by our method for the same latent code. We fine-tune StyleGAN2-ADA model with our vision-aided adversarial loss. On average we observe improved image quality with our method for the same latent code. }
    \lblfig{fig:appendix-train_vary_samples}
\end{figure*}

\begin{figure*}[!t]
    \centering
    \includegraphics[width=0.95\textwidth]{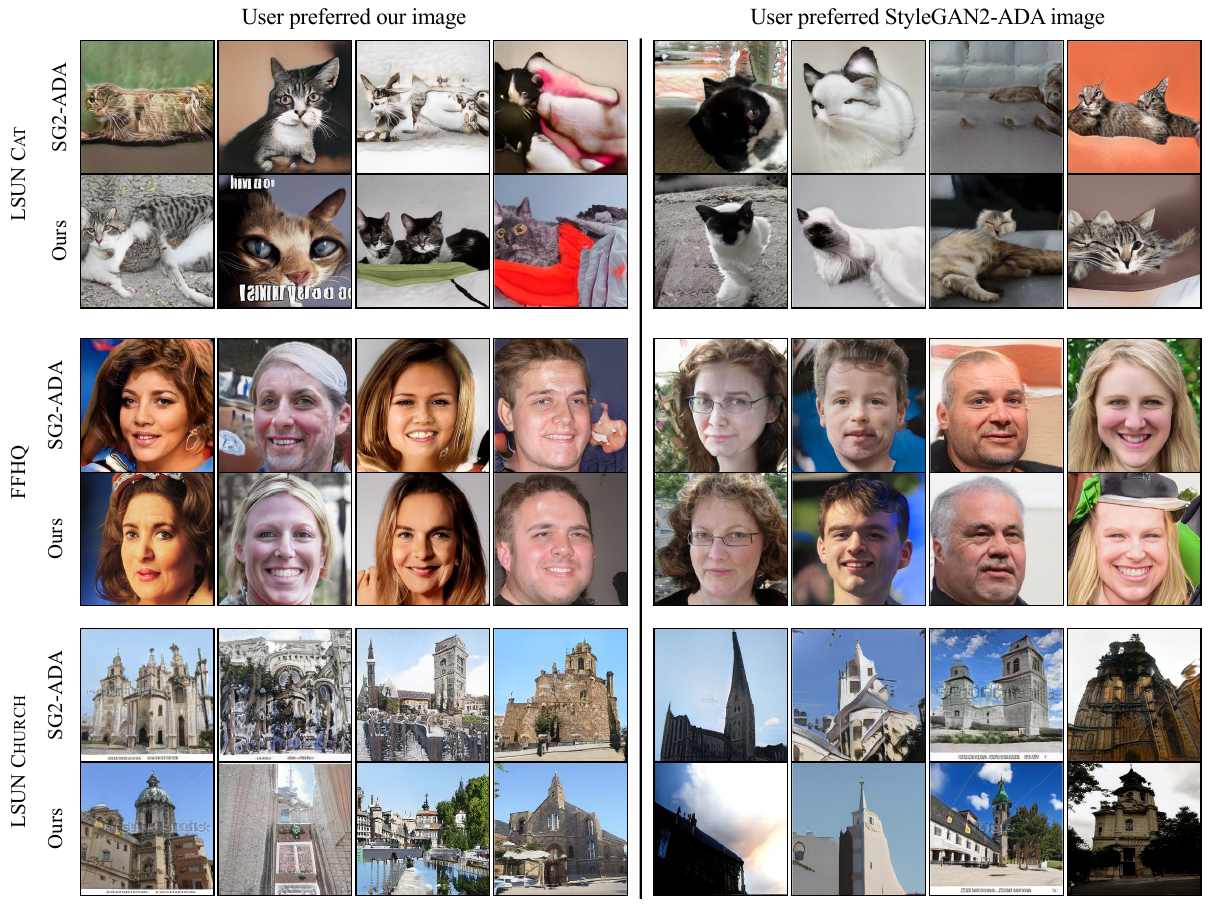}
    \caption{\textbf{Example images shown to users in the human preference study between our method (w/ ADA) and StyleGAN2-ADA} on \textsc{LSUN Cat}, \textsc{FFHQ}, and \textsc{LSUN Church} 1k training sample setting. \textit{Left:} example instances where images generated by our method is preferred by users. \textit{Right:} where images generated by StyleGAN2-ADA is preferred. For each dataset, top and bottom row show the two images generated by StyleGAN2-ADA and our method from the same random latent code and shown to the user.
    For \textsc{LSUN Cat}, \textsc{FFHQ}, and \textsc{LSUN Church} our method is preferred with $63.5\%$, $53.8\%$, and $60.5\%$ as mentioned in the main paper.
    }\lblfig{fig:appendix-human_study}
\end{figure*}

\begin{figure*}[!t]
    \centering
    \includegraphics[width=0.93\textwidth]{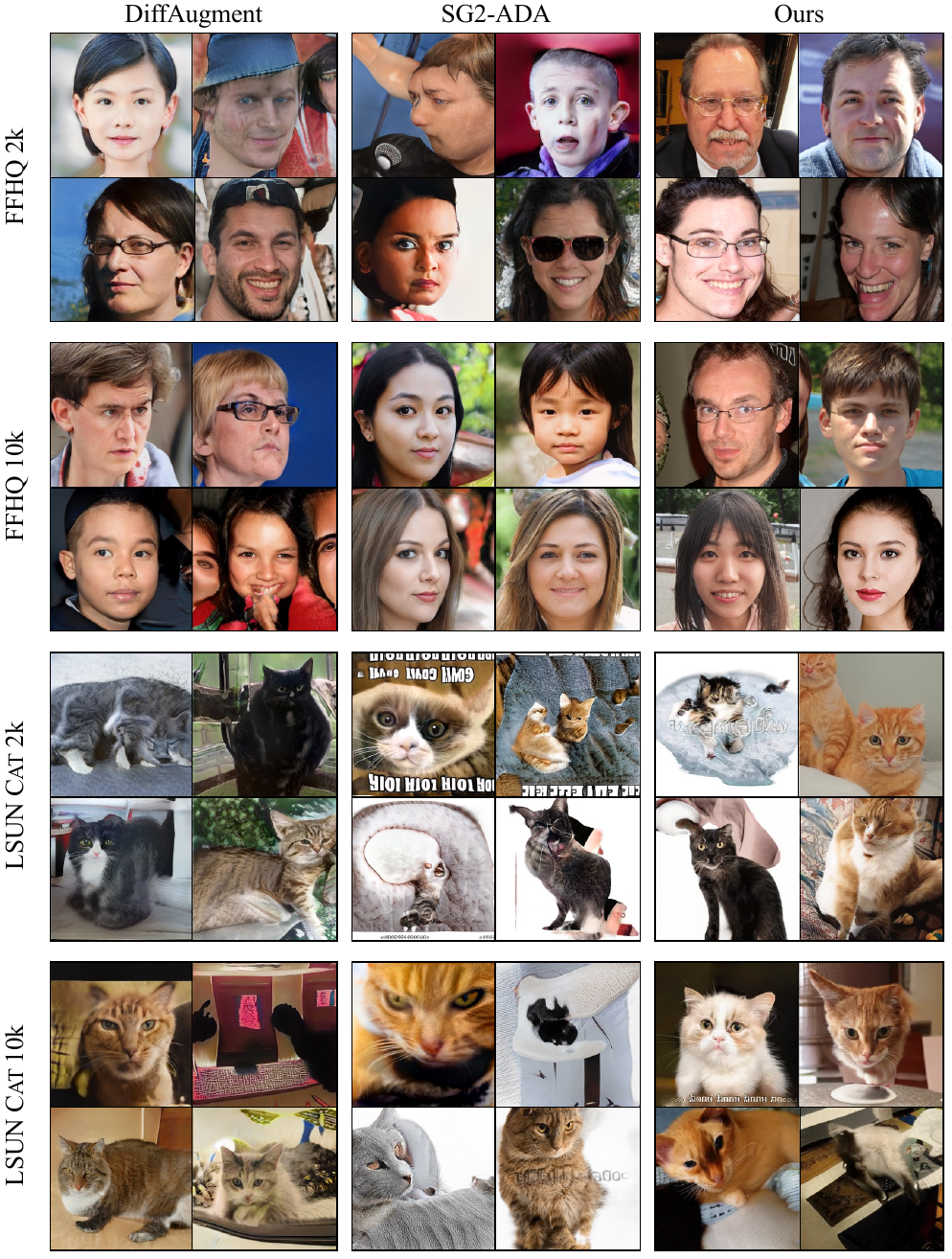}
    \caption{\textbf{Randomly generated samples} by DiffAugment~\cite{diffaug}, StyleGAN2-ADA~\cite{stylegan2ada} and Our method (w/ ADA) on 2k and 10k sample setting of \textsc{FFHQ} and \textsc{LSUN Cat}. 
    }\lblfig{fig:appendix-random_ffhq_cat}
      \vspace{-8pt}
\end{figure*}

\begin{figure*}[!t]
    \centering
    \includegraphics[width=0.93\textwidth]{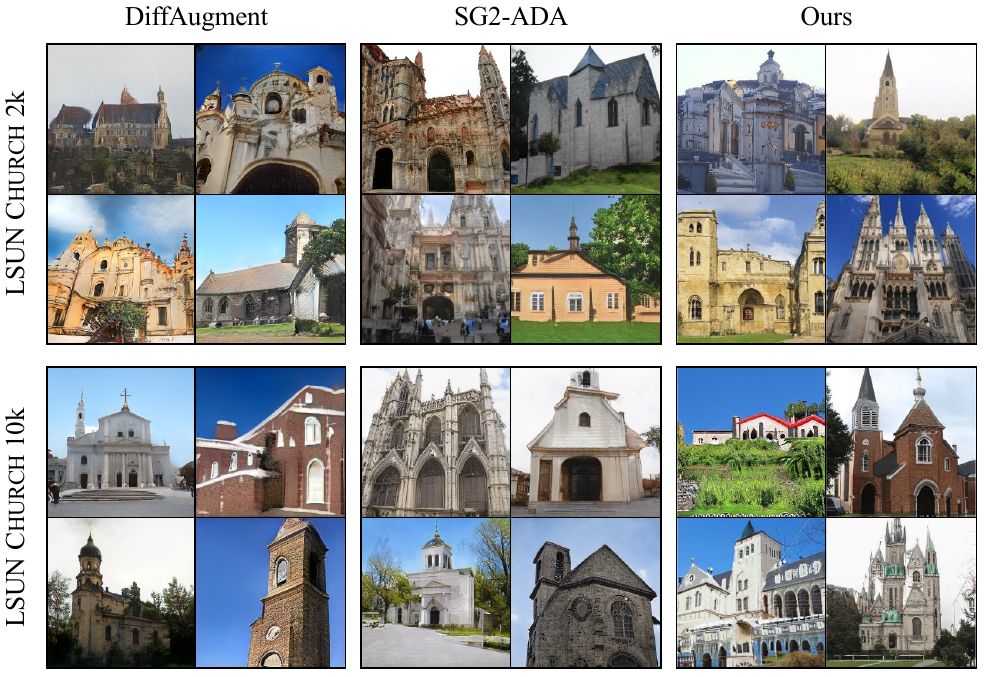}
    \caption{\textbf{Randomly generated samples} by DiffAugment~\cite{diffaug}, StyleGAN2-ADA~\cite{stylegan2ada} and Our method (w/ ADA) on \textsc{LSUN Church} 2k and 10k training sample setting. 
    }\lblfig{fig:appendix-random_ffhq_church}
      \vspace{-8pt}
\end{figure*}
\begin{figure*}[!t]
    \centering
    \includegraphics[width=0.9\textwidth]{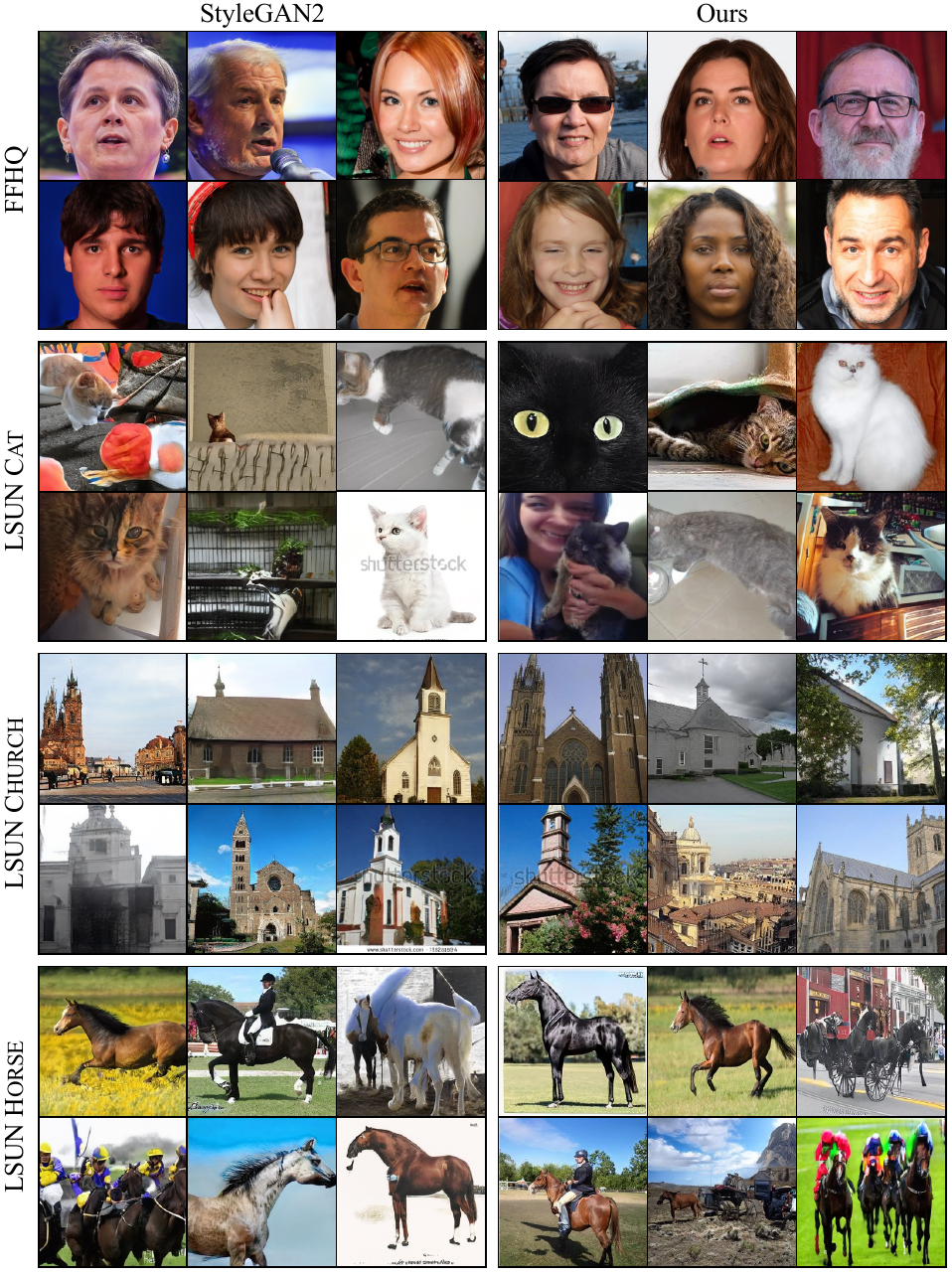}
    \caption{\textbf{Uncurated samples} generated by StyleGAN2~\cite{stylegan2} and Our method (w/ ADA) trained on full-dataset of \textsc{FFHQ}, \textsc{LSUN Cat}, \textsc{LSUN Church}, and \textsc{LSUN Horse}.
    }\lblfig{fig:appendix-random_ffhq_lsun_full}
      \vspace{-8pt}
\end{figure*}

\begin{figure*}[!t]
    \centering
    \includegraphics[width=0.93\textwidth]{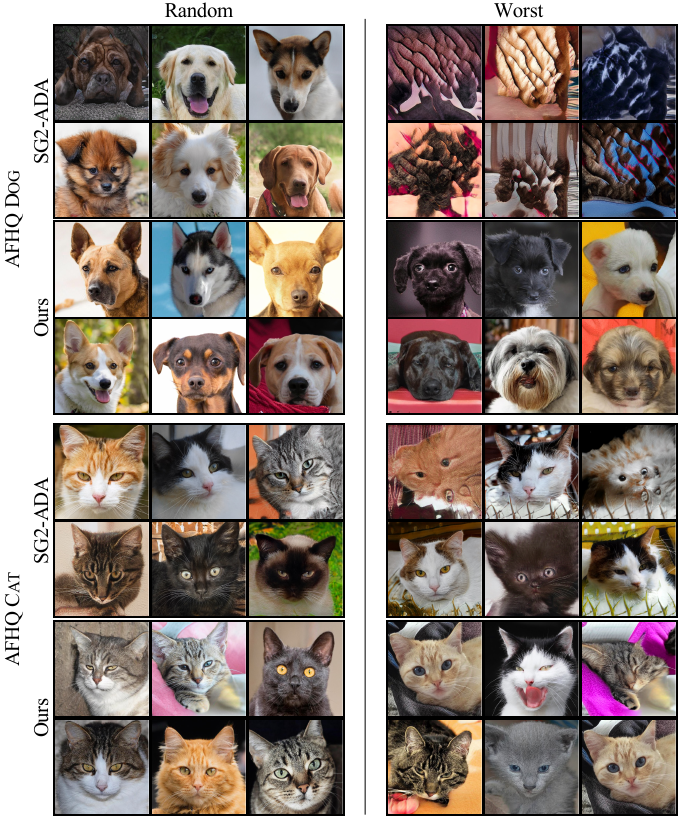}
    \caption{\textbf{Qualitative comparison of our method (w/ ADA) with StyleGAN2-ADA on \textsc{AFHQ Dog} and \textsc{AFHQ Cat}.} \textit{Left:} randomly generated samples for both methods. \textit{Right:} \textbf{Worst FID samples}. For both our model and StyleGAN2-ADA, we independently generate 5k samples and find the worst-case samples compared to real image distribution. We first fit a Gaussian model using the Inception~\cite{szegedy2016rethinking} feature space of real images. We then calculate the log-likelihood of each sample given this Gaussian prior and show the images with minimum log-likelihood (maximum Mahalanobis distance). Our method shows better image quality on average compared to StyleGAN2-ADA.
    }\lblfig{fig:appendix-afhq}
      \vspace{-8pt}
\end{figure*}
\begin{figure*}[!t]
    \centering
    \includegraphics[width=0.93\textwidth]{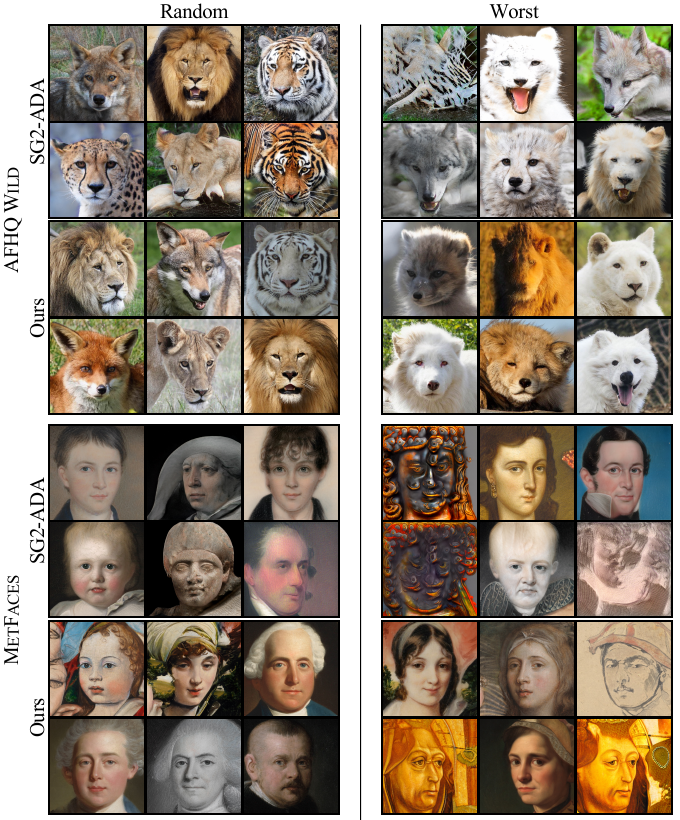}
    \caption{\textbf{Qualitative comparison of our method (w/ ADA) with StyleGAN2-ADA on \textsc{AFHQ Wild} and \textsc{MetFaces}} \textit{Left:} randomly generated samples for both methods. \textit{Right:} samples with maximum Mahalanobis distance as described in \reffig{fig:appendix-afhq}. 
    }\lblfig{fig:appendix-afhq2}
      \vspace{-8pt}
\end{figure*}

\begin{figure*}[!t]
    \centering
    \includegraphics[width=0.93\textwidth]{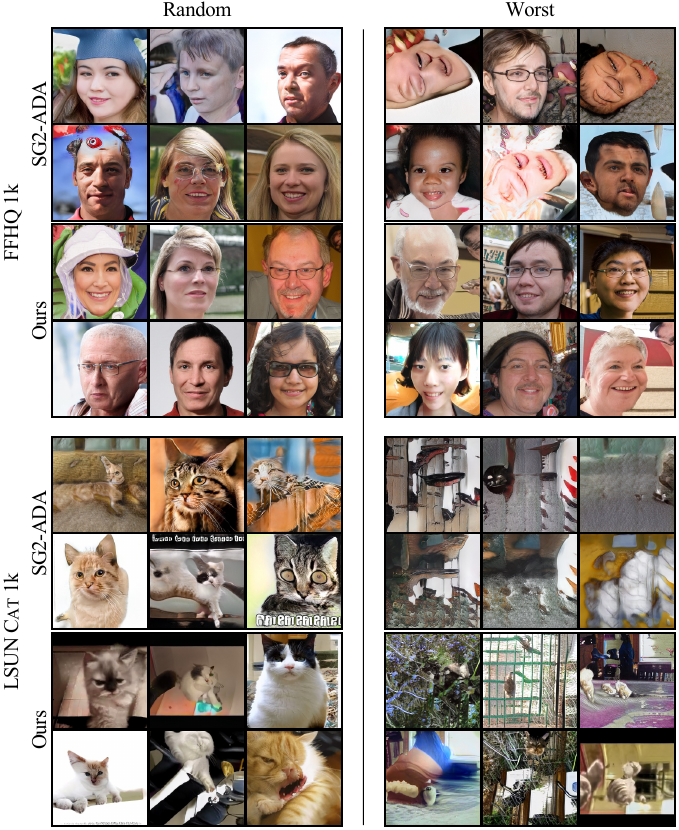}
    \caption{\textbf{Qualitative comparison of our method (w/ ADA) with StyleGAN2-ADA on \textsc{FFHQ} and \textsc{LSUN Cat} 1k training sample setting} \textit{Left:} randomly generated samples for both methods. \textit{Right:} samples with maximum Mahalanobis distance as described in \reffig{fig:appendix-afhq}. Our method to a large extent prevents generation of rotated images with extreme artifacts in case of \textsc{FFHQ} 1k training sample setting.
    }\lblfig{fig:appendix-afhq3}
      \vspace{-8pt}
\end{figure*}

\begin{figure*}[!t]
    \centering
    \includegraphics[width=0.93\textwidth]{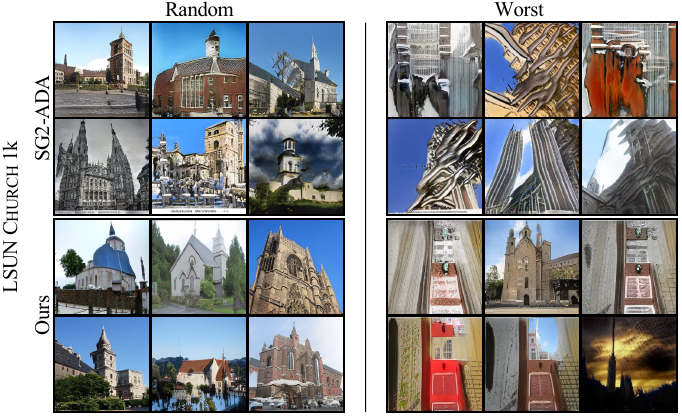}
    \caption{\textbf{Qualitative comparison of our method (w/ ADA) with StyleGAN2-ADA on \textsc{LSUN Church} 1k} \textit{Left:} randomly generated samples for both methods. \textit{Right:} samples with maximum Mahalanobis distance as described in \reffig{fig:appendix-afhq}. Our method has relatively better worst case samples compared to StyleGAN2-ADA.
    }\lblfig{fig:appendix-afhq4}
      \vspace{-8pt}
\end{figure*}

\begin{figure*}[!t]
    \centering
    \includegraphics[width=0.93\textwidth]{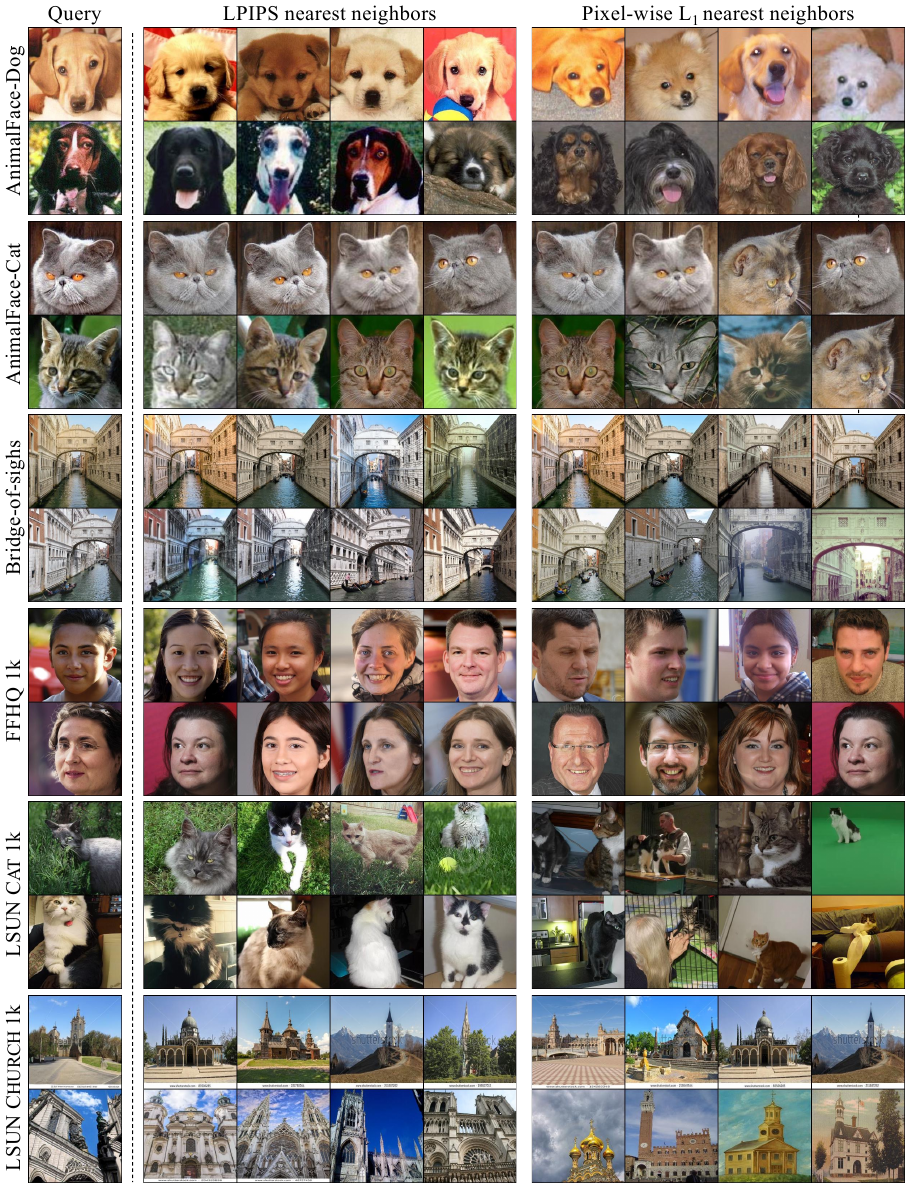}
    \caption{Nearest neighbor test on low-shot data settings. \textit{Left column}: generated images by our model. \textit{Middle column}: LPIPS based nearest neighbors from the training set. \textit{Right column}: pixel wise $L_1$ distance based nearest neighbors. We observe that the generated images are different from the training set. Thus our model is not simply memorizing the training set.
    }\lblfig{fig:appendix-nn-test}
      \vspace{-8pt}
\end{figure*}

\begin{figure*}[!t]
    \centering
    \includegraphics[width=0.93\textwidth]{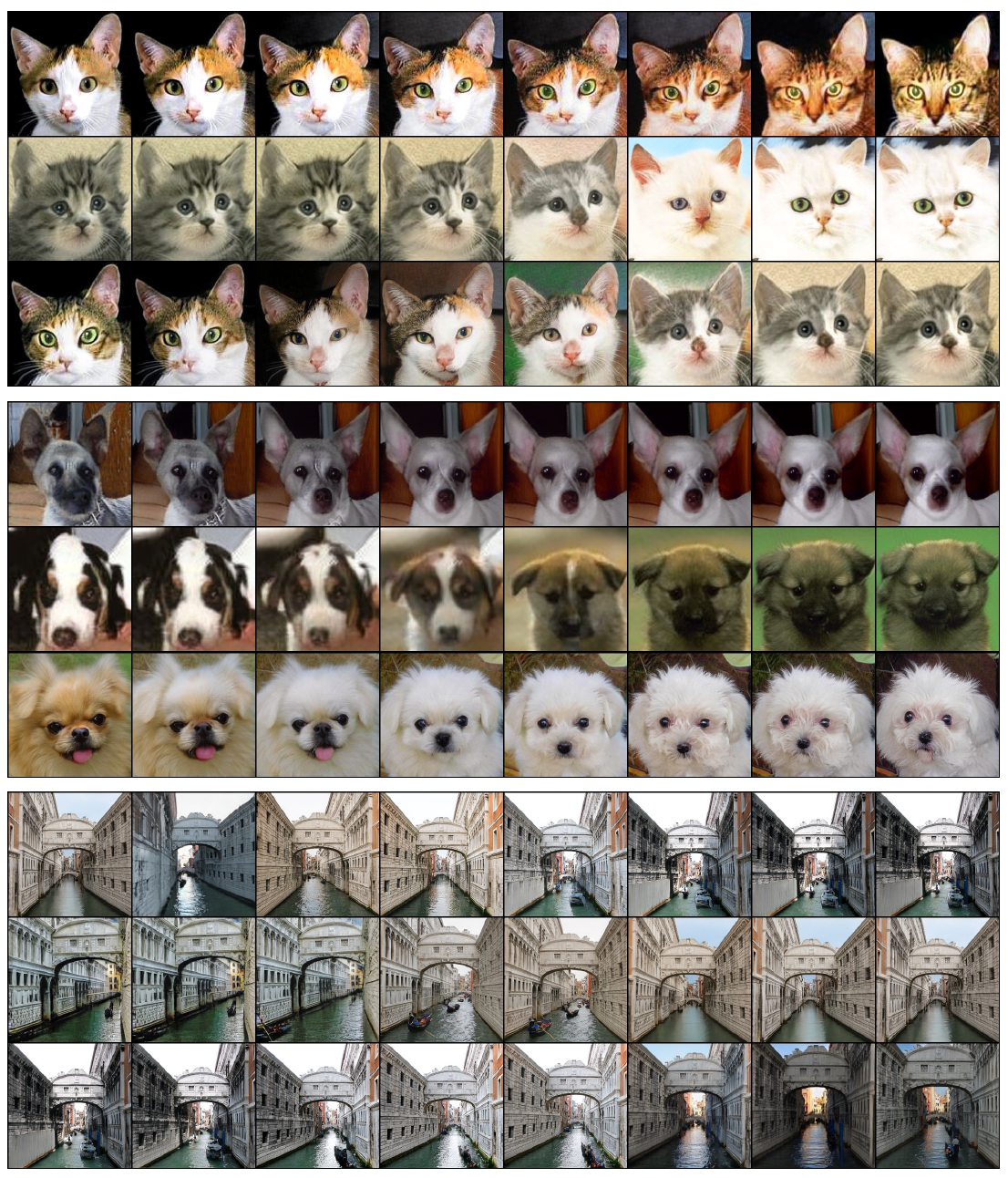}
    \caption{\textbf{Latent interpolation results} of models trained with our method on AnimalFace  Cat (169 images), AnimalFace Dog (389 images)~\cite{si2011learning} and 100-shot Bridge-of-Sighs~\cite{diffaug} datasets. The smooth interpolation suggests that there is probably little overfitting in the trained generator.
    }\lblfig{fig:appendix-low_shot}
      \vspace{-8pt}
\end{figure*}

\begin{figure*}[t]
    \centering
    \includegraphics[width=\textwidth]{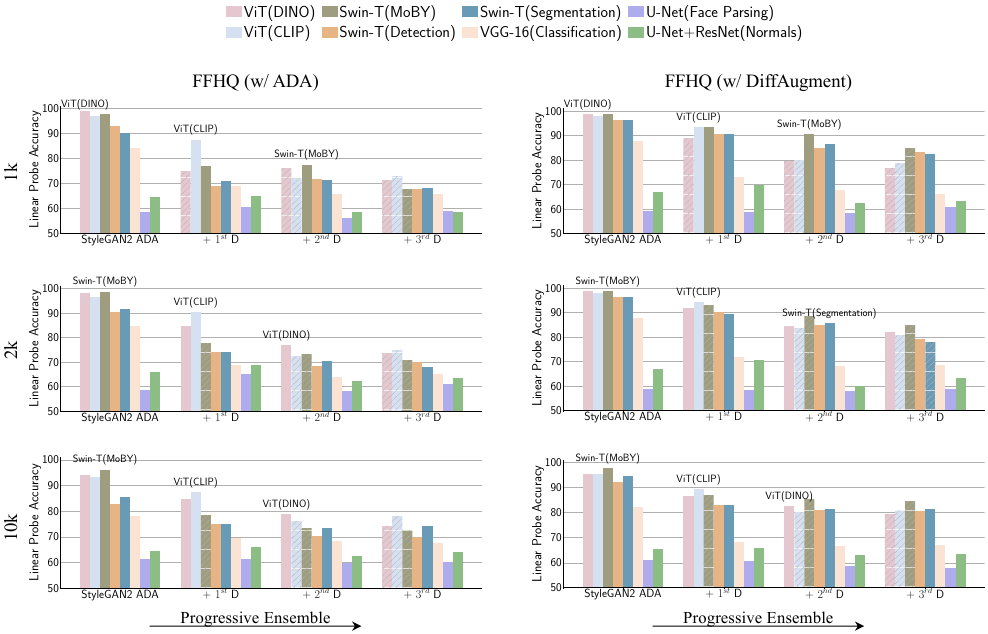}
    \caption{\textbf{Linear probe accuracy of off-the-shelf models during our K-progressive ensemble training} on \textsc{FFHQ} with different training sample setting for both ADA and DiffAugment. The selected model at each stage is annotated at the top of the bar-plot. As we include more vision-aided discriminators during GAN training, linear probe accuracy of the pretrained models decreases. }
    \lblfig{fig:appendix-linear_acc_all1}
    \vspace{-10pt}
\end{figure*}

\begin{figure*}[t]
    \centering
    \includegraphics[width=\textwidth]{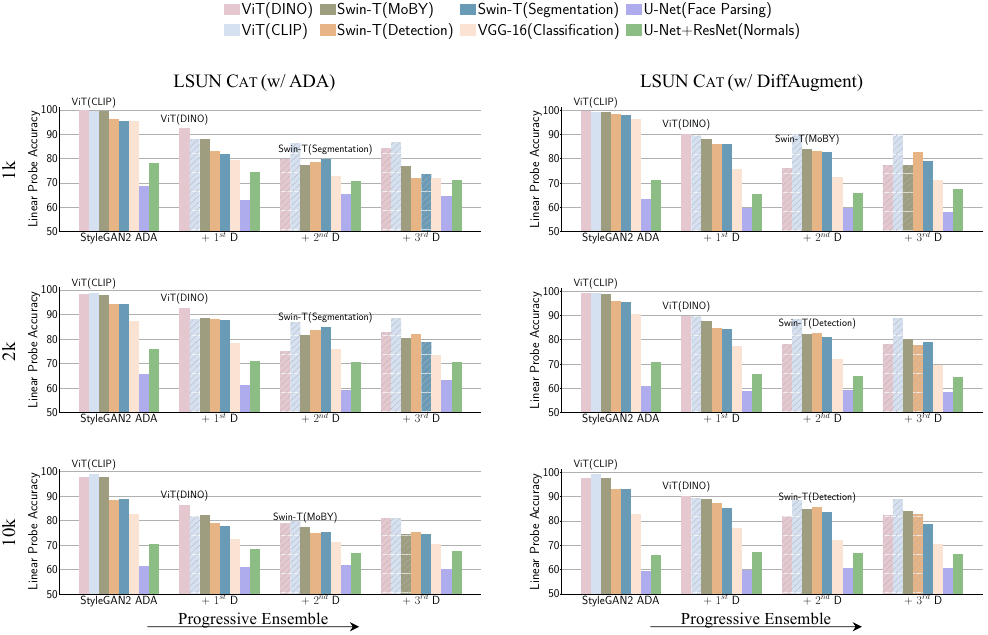}
    \caption{\textbf{Linear probe accuracy of off-the-shelf models during our K-progressive ensemble training} on \textsc{LSUN Cat} with different training sample setting for both ADA and DiffAugment. The selected model at each stage is annotated at the top of the bar-plot. As we include more vision-aided discriminators during GAN training, linear probe accuracy of the pretrained models decreases.}
    \lblfig{fig:appendix-linear_acc_all2}
    \vspace{-10pt}
\end{figure*}

\begin{figure*}[t]
    \centering
    \includegraphics[width=\textwidth]{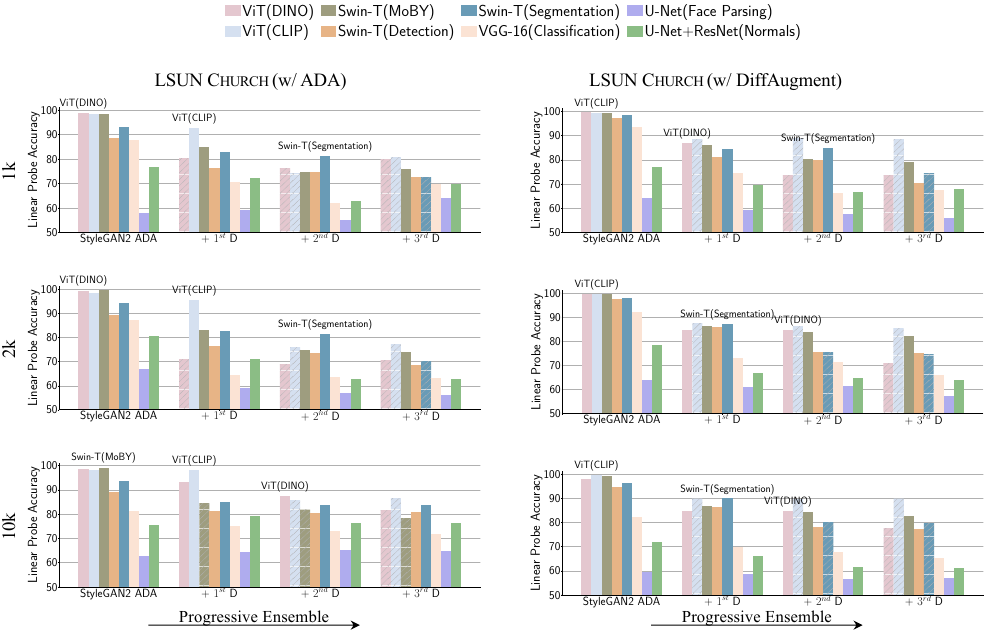}
    \caption{\textbf{Linear probe accuracy of off-the-shelf models during our K-progressive ensemble training} on \textsc{LSUN Church} with different training sample setting for both ADA and DiffAugment. The selected model at each stage is annotated at the top of the bar-plot. As we include more vision-aided discriminators during GAN training, linear probe accuracy of the pretrained models decreases. }
    \lblfig{fig:appendix-linear_acc_all3}
    \vspace{-10pt}
\end{figure*}

\begin{figure*}[t]
    \centering
    \includegraphics[width=\textwidth]{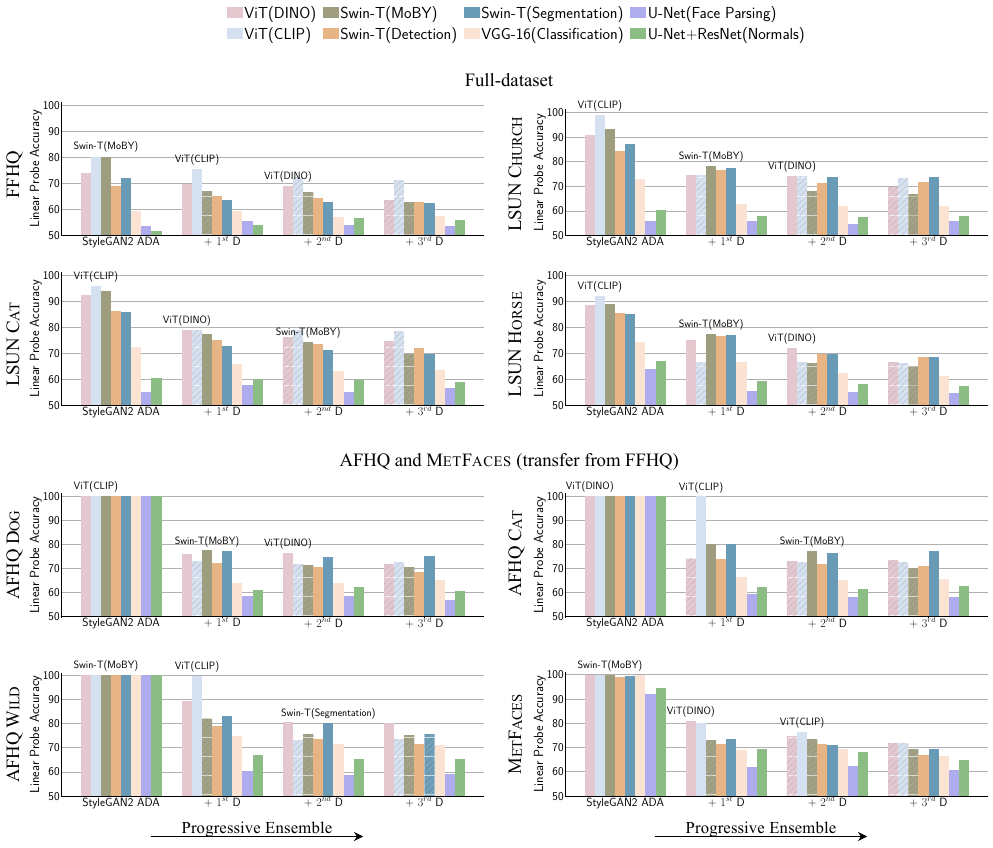}
    \caption{\textbf{Linear probe accuracy of off-the-shelf models during our K-progressive ensemble training} on full-dataset of \textsc{FFHQ}, \textsc{LSUN} categories and \textsc{AFHQ}, \textsc{MetFaces} (transfer from FFHQ trained generator). In case of transfer from FFHQ, linear probe accuracy is $100\%$ at the start as human faces and AFHQ categories have a significant domain gap. The selected model at each stage is annotated at the top of the bar-plot. As we include more vision-aided discriminators during GAN training, linear probe accuracy of the pretrained models decreases.}
    \lblfig{fig:appendix-linear_acc_all4}
    \vspace{-10pt}
\end{figure*}

\begin{figure*}[t]
    \centering
    \includegraphics[width=\textwidth]{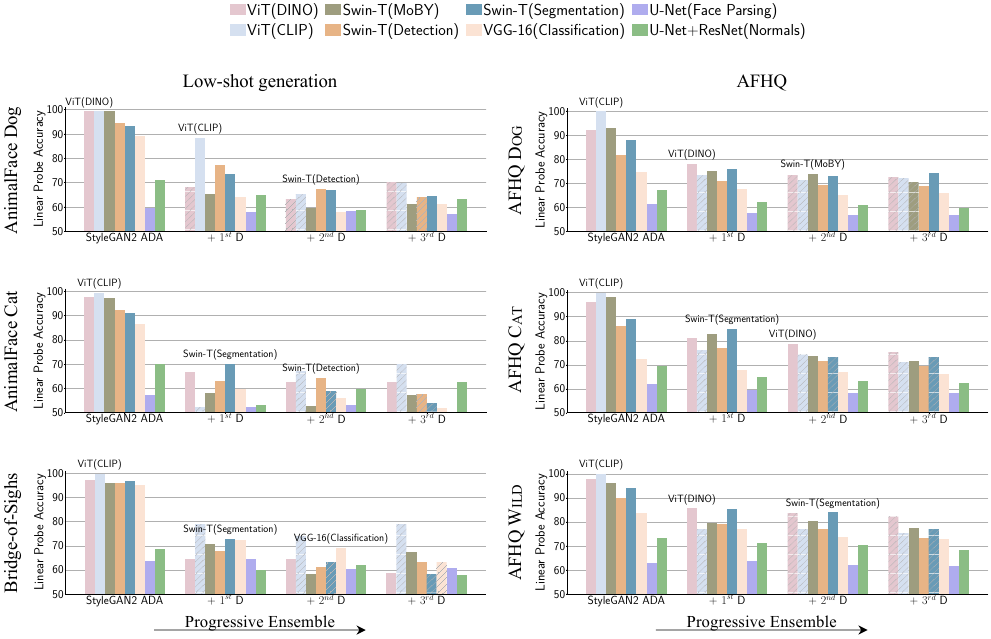}
    \caption{\textbf{Linear probe accuracy of off-the-shelf models during our K-progressive ensemble training} on AnimalFace Cat, Dog~\cite{si2011learning} and 100-shot Bridge-of-Sighs~\cite{diffaug} low-shot datasets, and \textsc{AFHQ} categories. The selected model at each stage is annotated at the top of the bar-plot. As we include more vision-aided discriminators during GAN training, linear probe accuracy of the pretrained models decreases. }
    \lblfig{fig:appendix-linear_acc_all5}
    \vspace{-10pt}
\end{figure*}

\clearpage

\end{document}